\documentclass{eitnlp}
\usepackage{xurl}
\usepackage{array,colortbl,wrapfig}
\usepackage{placeins}

\newtheorem{proposition}{Proposition}
\newcommand{\method}{\textsc{DIAL-OPD}}
\DeclareMathSizes{7}{7}{5}{5}
\definecolor{methodblue}{RGB}{234,241,248}
\definecolor{studentband}{RGB}{235,232,241}
\definecolor{legendblue}{RGB}{225,238,247}
\definecolor{legendpink}{RGB}{245,225,237}
\definecolor{legendgray}{RGB}{235,232,238}
\DeclareRobustCommand{\legendword}[2]{{\setlength{\fboxsep}{1pt}\colorbox{#1}{\strut #2}}}

\newsavebox{\resultstablebox}

\let\eitOriginalUsebox\usebox
\renewcommand{\usebox}[1]{%
  \ifx#1\resultstablebox
    \ifdim\wd#1>\linewidth
      \resizebox{\linewidth}{!}{\eitOriginalUsebox{#1}}%
    \else
      \eitOriginalUsebox{#1}%
    \fi
  \else
    \eitOriginalUsebox{#1}%
  \fi
}

\usepackage{etoolbox}
\makeatletter
\patchcmd{\maketitle}
  {\vspace{4pt}{\small\eit@contact\par}}
  {\vspace{-\parskip}\vspace{1pt}{\small\eit@contact\par}}
  {}{\PackageError{eitnlp-setup}{Contact spacing patch failed}{Check the template maketitle definition.}}
\makeatother

\title{DIAL-OPD: Learning More from Fewer Tokens in On-Policy Distillation}
\runningtitle{DIAL-OPD}
\author{%
  \textbf{Anhao Zhao}\textsuperscript{1,2}\qquad
  \textbf{Haoran Xin}\textsuperscript{3}\qquad
  \textbf{Junlong Tong}\textsuperscript{1,4}\qquad
  \textbf{Yingqi Fan}\textsuperscript{5}\\[4pt]
  \textbf{Xuan Lu}\textsuperscript{1,4}\qquad
  \textbf{Ping Nie}\textsuperscript{6}\qquad
  \textbf{Wenjie Li}\textsuperscript{2}\qquad
  \textbf{Xiaoyu Shen}\textsuperscript{1\textdagger}}
\affiliation{%
  \textsuperscript{1}EIT-NLP Lab, Eastern Institute of Technology, Ningbo\\
  \textsuperscript{2}The Hong Kong Polytechnic University \quad
  \textsuperscript{3}HKUST (GZ)\\
  \textsuperscript{4}Shanghai Jiaotong University\quad
  \textsuperscript{5}University of Hong Kong\quad
  \textsuperscript{6}University of Waterloo}
\authornotes{\textsuperscript{\textdagger}Corresponding author.}
\contact{\texttt{anhao.zhao@connect.polyu.hk}\qquad\texttt{xyshen@eitech.edu.cn}}
\paperlinks{%
  \href{https://github.com/EIT-NLP/DIAL-OPD}{%
    \textcolor{eitblue}{\faGithub\enspace
    \texttt{EIT-NLP/DIAL-OPD}}}}
\usepackage{fontawesome5}
\begin{document}
\maketitle
\suppressfloats[t]
\begin{abstract}
On-policy distillation (OPD) supervises student-generated trajectories with token-level teacher signals, and its sampled-token variant has become the practical default by avoiding the cost of full-vocabulary probabilities. Yet we find that supervising more tokens is not better: because the student's capacity is finite, training on a carefully chosen subset can significantly outperform full-token OPD. Effective OPD therefore hinges not on how much supervision to provide, but on which tokens deserve it. Existing disagreement-based criteria, however, are scale-blind: the log-ratio reward treats a token the teacher strongly endorses and one that both models assign negligible probability as equally informative, causing supervision to concentrate on \emph{low-low} tokens that hinder learning. We propose \textbf{DIAL-OPD}, a token-selection method that bridges log-probability and probability spaces by weighting reward magnitude with the logarithmic mean of teacher and student probabilities. The parameter $\beta$ controls this weighting, and the highest-scoring tokens are retained for training. 
We evaluate DIAL-OPD against 9 baselines across 4 teacher--student pairs and 7 mathematical reasoning benchmarks, showing that retaining only 40\% of tokens, it outperforms Vanilla OPD and its full-token variants by up to \textbf{5.25} percentage points in mean accuracy, while doubling Vanilla OPD's AIME25 Pass@16 from 13.33\% to 26.67\%. It also outperforms the strongest token-selection baseline at matched retention ratios, with up to an \textbf{18}\% relative improvement in mean accuracy.
Even with a 4B teacher, DIAL-OPD surpasses the strongest full-token baseline using an 8B teacher at both student scales, \textbf{showing that effective supervision allocation can outweigh teacher scaling.}
Further analysis shows that moderate $\beta$ best balances suppressing low-low tokens against preserving useful disagreements, while token-level evidence reveals that DIAL-OPD removes high-reward tokens with limited reasoning value without sacrificing supervision critical to reasoning correctness.
\end{abstract}

\section{Introduction}
\label{sec:introduction}

On-policy distillation (OPD) has become a standard component of LLM post-training~\citep{qwen2025qwen3,lu2025opd,deepseek2026v41flash}. By supervising student-generated trajectories, OPD reduces the exposure bias of fixed-response distillation~\citep{agarwal2024gkd,gu2024minillm} while providing dense token-level guidance beyond the sparse outcome rewards used in reinforcement learning~\citep{deepseek2025r1,zhao2026decoupling}. A straightforward formulation is to minimize the reverse KL divergence between the full teacher and student distributions at every generated token. However, obtaining and transmitting vocabulary-wide teacher probabilities is prohibitively expensive in both computation and memory~\citep{slime2026opd,helwig2026verlopd}. As a result, the prevailing practical formulation is \emph{sampled-token OPD}, which uses the token sampled by the student to construct an unbiased single-sample Monte Carlo estimate of the reverse KL and optimizes the student with a token-level log-ratio reward~\citep{kimi2026k3,glm2026glm5,xiaomi2026mimov2flash}.

\begin{figure}[t]
\centering
\begin{minipage}[t]{0.49\linewidth}
\centering
\includegraphics[width=\linewidth]{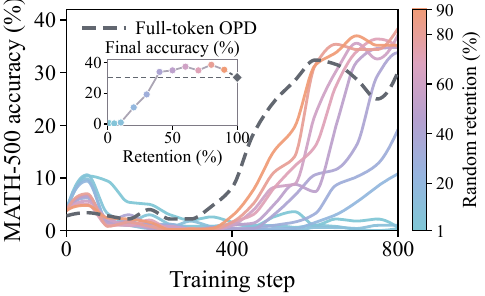}\par
\vspace{-10pt}
\caption{\small\textbf{Supervision density has a non-monotonic effect.} Random selection (1--90\%) vs. full-token OPD (100\%), 0.6B student; inset: final accuracy.}
\label{fig:supervision-density}
\end{minipage}
\hfill
\begin{minipage}[t]{0.49\linewidth}
\centering
\includegraphics[width=\linewidth]{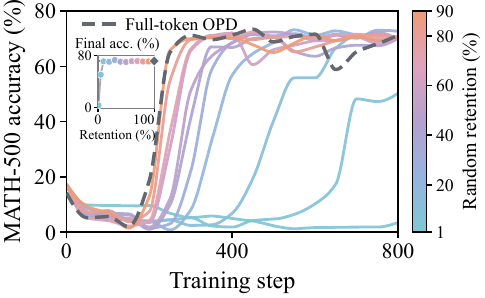}\par
\vspace{-10pt}
\caption{\small\textbf{The non-monotonic effect persists without a teacher--student size gap.} Random selection vs. full-token OPD, 4B student; inset: final accuracy.}
\label{fig:supervision-density-matched}
\end{minipage}
\end{figure}

While sampled-token OPD retains the dense, token-level nature of OPD, its supervision is fundamentally noisier: rather than comparing the full distributions, each token provides only a single-sample estimate of the teacher--student discrepancy. This raises a basic question: \emph{if the key advantage of OPD is dense token-level supervision, should we necessarily supervise every token even under sampled-token OPD?} We investigate this question with Qwen3-4B as the teacher and Qwen3-0.6B as the student. Surprisingly, randomly retaining only a subset of tokens consistently outperforms full-token Vanilla OPD across all tested retention ratios from 40\% to 90\%, with 80\% retention improving MATH-500 accuracy by \textbf{8.2} percentage points (Figure~\ref{fig:supervision-density}). This advantage disappears when the student is enlarged to match the 4B teacher (Figure~\ref{fig:supervision-density-matched}), indicating that the benefit is closely tied to student capacity. A limited-capacity student cannot absorb all noisy token-level signals equally well; forcing it to learn from every token can therefore dilute useful supervision with signals that are less informative or harder to learn. These results suggest that, in sampled-token OPD, \emph{more supervision is not necessarily better}: the key is to spend limited learning capacity on the tokens that provide the most useful signals.

This observation naturally leads to the next question: \emph{which tokens should be selected?} Random selection treats all tokens equally and therefore ignores differences in their learning value. We first consider teacher--student disagreement as a natural selection criterion~\citep{xu2026tip,shao2026tokenanalysis,yu2026tide}. In sampled-token OPD, the log-ratio reward compares the probabilities that the teacher and student assign to the sampled token, making its magnitude a natural measure of their disagreement. We find that selecting tokens with the largest absolute rewards outperforms both selecting those with the smallest rewards and random selection across various retention ratios (Figure~\ref{fig:supervision-selection}). These results support larger reward magnitudes as a proxy for greater token learning value. However, this strategy exhibits prolonged stagnation early in training, and its final performance does not consistently exceed full-token Vanilla OPD.
Examining the log-ratio reward reveals a key blind spot: \textbf{it measures relative disagreement but ignores absolute probability scale.} For example, student--teacher probabilities of $(0.08,0.8)$ and $(10^{-4},10^{-3})$ yield the same reward, although the former reflects strong teacher support and the latter negligible probability under both models. We call tokens in the latter regime \emph{low-low tokens}. Prior work identifies reliability issues in low-probability generation tails~\citep{holtzman2020degeneration} and shows that suppressing rewards where both models assign low probability improves OPD training~\citep{zhao2026poweropd}, motivating closer scrutiny of these tokens. We find that (i) when selecting tokens with the largest absolute rewards at 10\% retention, low-low tokens account for only \textbf{20.6}\% of selected tokens but \textbf{49.8}\% of total absolute reward (Figure~\ref{fig:low-low-dynamics}); (ii) replacing selected low-low tokens with non-low-low tokens at the same retention ratio mitigates early training stagnation and improves final accuracy across various retention ratios (Figure~\ref{fig:low-low-replacement}). Selecting tokens solely by reward magnitude can therefore concentrate supervision on tokens that hinder learning. \emph{To address this blind spot, token selection should consider both the relative disagreement captured by reward magnitude and the absolute probability scale at which it occurs.}

Building on these findings, we propose \textbf{DIAL-OPD}, a token-selection method that bridges log-probability and probability spaces. We use the logarithmic mean of the probabilities assigned to each token by the teacher and student to incorporate probability scale into reward-based selection. A tunable parameter $\beta$ controls the strength of probability weighting, reducing the priority of low-low tokens for $\beta>0$. The resulting criterion provides a continuous dial: $\beta=0$ recovers selection based solely on reward magnitude, $\beta=1$ corresponds to selection by the absolute probability gap, and intermediate values geometrically interpolate between the two, while $\beta>1$ further strengthens probability weighting. DIAL-OPD selects the highest-scoring tokens within each response at a specified retention ratio and applies the original OPD loss only to the selected tokens.

Experiments show that incorporating probability scale mitigates early training stagnation and improves final accuracy, with extensive $\beta$ sweeps favoring moderate weighting. We evaluate DIAL-OPD against 9 baselines across 4 teacher--student pairs and 7 mathematical reasoning benchmarks. Retaining only \textbf{40}\% of tokens, DIAL-OPD surpasses Vanilla OPD and the strongest OPD variant, which train on all response tokens, in mean accuracy across all settings, by up to \textbf{5.25} and \textbf{1.24} percentage points, respectively. Notably, it doubles Vanilla OPD's AIME25 Pass@16 from \textbf{13.33}\% to \textbf{26.67}\%. At \textbf{20}\% retention, DIAL-OPD exceeds the strongest token-selection baseline at the same ratio by up to \textbf{5.23} points in mean accuracy and \textbf{10.33} points on GSM8K Avg@4. With a 4B teacher, it also surpasses the strongest full-token reference using an 8B teacher by up to \textbf{1.76} mean accuracy points, \emph{demonstrating that effective supervision allocation can outweigh teacher scaling.} Selection analyses suggest that moderate weighting suppresses low-low tokens, whereas excessive weighting may discard useful disagreements. Token-level studies reveal that DIAL-OPD filters high-reward tokens with limited reasoning value while preserving supervision essential to reasoning correctness.

\section{Preliminaries}

\label{sec:preliminaries}
\label{sec:prelim-opd}
\label{sec:prelim-reward}

On-policy distillation (OPD) aligns a student policy $\pi_\theta$ with a fixed teacher $\pi_T$ on student-generated responses~\citep{agarwal2024gkd,gu2024minillm}. For $x\sim\mathcal D$, $y\sim\pi_\theta(\cdot\mid x)$, and $c_t=(x,y_{<t})$, OPD minimizes the reverse KL between their next-token distributions over vocabulary $\mathcal V$:
\begin{equation}
\mathcal L_{\mathrm{OPD}}(\theta)
=\mathbb E_{x,y}
\left[
\sum_{t=1}^{|y|}
D_{\mathrm{KL}}\!\left(
\pi_\theta(\cdot\mid c_t)\,\|\,\pi_T(\cdot\mid c_t)
\right)
\right],
\quad
D_{\mathrm{KL}}(P\|Q)
=\sum_{v\in\mathcal V}P(v)\log\frac{P(v)}{Q(v)}.
\label{eq:opd-objective}
\end{equation}

Transferring and storing full-vocabulary teacher distributions incurs substantial communication and memory costs; sampled-token OPD reduces these costs by querying only the teacher log-probabilities of student-generated tokens~\citep{slime2026opd,helwig2026verlopd}. With the rollout context distribution held fixed during differentiation, the reverse-KL descent direction takes the policy-gradient reinforcement learning (RL) form~\citep{williams1992reinforce,sutton1999policygradient}:
\begin{equation}
g_{\mathrm{OPD}}(\theta)
=\mathbb E_{x\sim\mathcal D,\,y\sim\pi_\theta(\cdot\mid x)}
\Biggl[
\sum_{t=1}^{|y|}
\underbrace{
\log\frac{\pi_T(y_t\mid c_t)}{\pi_\theta(y_t\mid c_t)}
}_{r_t^{\mathrm{OPD}}}
\nabla_\theta\log\pi_\theta(y_t\mid c_t)
\Biggr].
\label{eq:opd-update}
\end{equation}
Student rollouts provide an unbiased Monte Carlo estimate of $g_{\mathrm{OPD}}$, with reward $r_t^{\mathrm{OPD}}$ treated as a stop-gradient weight (\citealp{lu2025opd,ko2026reopold,zhao2026poweropd}; derivation in Appendix~\ref{app:opd-update-derivation}). This formulation is adopted by recent open-source LLMs~\citep{glm2026glm5,kimi2026k3,xiaomi2026mimov2flash} and is the focus of this work.

\section{Rethinking Supervision Allocation in On-Policy Distillation}
\label{sec:analysis}

\subsection{More Supervision Does Not Imply Better Learning}
\label{sec:analysis-supervision}

Dense supervision is a central advantage of OPD, providing teacher guidance at every generated token~\citep{lu2025opd}. To test the implicit expectation that \textbf{denser supervision improves learning}, we conduct a controlled experiment comparing full-token training with uniform random token selection under a fixed update budget.

\paragraph{Experimental setup.}
We train Qwen3-0.6B with a Qwen3-4B teacher~\citep{qwen2025qwen3} on DeepScaleR~\citep{deepscaler2025}. Under a fixed update budget, we compare uniform random selection within each response ($\rho\in\{1\%,5\%,10\%,20\%,\ldots,90\%\}$) with full-token Vanilla OPD ($\rho=100\%$). All tokens remain in context; the OPD loss is averaged over selected tokens in each batch. We track MATH-500 accuracy throughout training~\citep{hendrycks2021math,lightman2023verify}; implementation and evaluation details appear in Appendix~\ref{app:supervision-density}.

\paragraph{Results.}
Surprisingly, random token selection outperforms full-token Vanilla OPD at all tested retention ratios from 40\% to 90\%, reaching 38.4\% accuracy at 80\% versus 30.2\%, a gain of \textbf{8.2} percentage points (Figure~\ref{fig:supervision-density}). However, this benefit may reflect the smaller student's limited capacity to utilize dense supervision from a larger teacher. To test this explanation, we use a Qwen3-4B student with the same teacher. Figure~\ref{fig:supervision-density-matched} shows that (i) 10\% retention nearly matches full-token Vanilla OPD (70.6\% vs.\ 71.2\%); and (ii) 30\% retention outperforms it (72.6\%). Together, these results show that \textbf{more supervision does not necessarily improve learning}, even at equal teacher--student model size. \emph{Since random selection already improves learning, how can we identify the token subset that maximizes learning performance under a fixed update budget?}

\subsection{Which Tokens Are Worth Learning From?}
\label{sec:analysis-reward}
\begin{figure}[t]
\centering
\begin{minipage}[t]{0.32\linewidth}
\centering
\includegraphics[width=\linewidth]{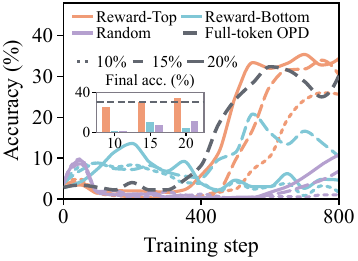}\par
\vspace{-10pt}
\caption{\small Reward-based selection; inset: final accuracy.}
\label{fig:supervision-selection}
\end{minipage}
\hfill
\begin{minipage}[t]{0.32\linewidth}
\centering
\includegraphics[width=\linewidth]{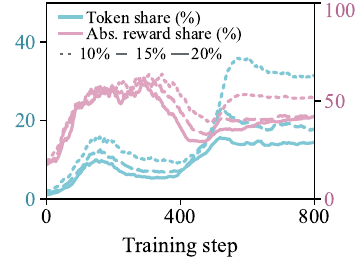}\par
\vspace{-10pt}
\caption{\small Low-low token and absolute reward shares (Reward-Top).}
\label{fig:low-low-dynamics}
\end{minipage}
\hfill
\begin{minipage}[t]{0.32\linewidth}
\centering
\includegraphics[width=\linewidth]{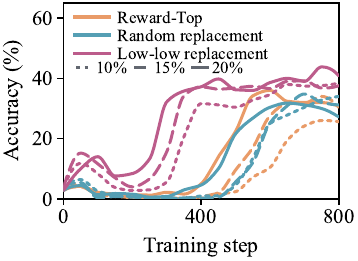}\par
\vspace{-10pt}
\caption{\small\textbf{Low-low replacement improves learning.}}
\label{fig:low-low-replacement}
\end{minipage}
\end{figure}

Random selection improves learning without distinguishing token value, motivating a more principled criterion. Teacher--student disagreement offers a natural starting point: recent work improves OPD by using disagreement or token update impact to guide selection~\citep{xu2026tip,shao2026tokenanalysis,yu2026tide}. Full-vocabulary OPD measures disagreement by KL divergence; sampled-token OPD uses a log-ratio reward that weights each token's policy-gradient term (Equation~\ref{eq:opd-update}). We therefore examine whether reward magnitude identifies tokens worth prioritizing.

\paragraph{Experimental setup.}
At 10\%, 15\%, and 20\% retention per response, we compare \emph{Reward-Top} (largest absolute rewards) with (i) \emph{Reward-Bottom} (smallest), testing whether larger rewards are more useful; and (ii) \emph{Random} (uniform sampling), testing whether reward-based selection outperforms uninformed selection. All methods use matched retention ratios, the original OPD loss, and the training and evaluation settings of Section~\ref{sec:analysis-supervision}.

\paragraph{Results.}
Reward-Top outperforms Reward-Bottom and Random at 10\%, 15\%, and 20\% retention, reaching 25.6\%, 30.4\%, and 34.4\% accuracy, respectively (Figure~\ref{fig:supervision-selection}). However, (i) all three runs show little improvement during the first 350 training steps, indicating that large rewards do not readily translate into learning gains; and (ii) Reward-Top does not consistently outperform full-token Vanilla OPD. Thus, reward magnitude helps distinguish token learning value but is insufficient for reliable selection. \emph{These limitations raise a question: what information does reward magnitude miss when assessing token learning value?}

\subsection{Low--Low Supervision Can Hinder Learning}
\label{sec:analysis-low-low}

To understand these limitations, we examine the log-ratio reward: it captures relative disagreement but ignores absolute probability scale. Student--teacher probabilities of $(0.08,0.8)$ and $(10^{-4},10^{-3})$ both yield $\log 10$, despite strong teacher support in the former and negligible probabilities under both models in the latter, which we term \emph{low-low tokens}. Prior work identifies unreliable low-probability generation tails~\citep{holtzman2020degeneration}, while PowerOPD improves training stability and performance by suppressing low-low rewards~\citep{zhao2026poweropd}. Accordingly, we test (i) whether Reward-Top overemphasizes low-low tokens and (ii) whether training on them hinders learning.

\paragraph{Experimental setup.}
Following Section~\ref{sec:analysis-reward}, we study Reward-Top at 10\%, 15\%, and 20\% retention, defining low-low tokens by student and teacher probabilities both at most 1\%. (i) To quantify their concentration, we track their shares of selected tokens and total absolute reward throughout training. (ii) To test whether they hinder learning, we replace selected low-low tokens with the highest-reward unselected non-low-low tokens (\emph{Low-low replacement}). As a control, \emph{Random replacement} removes the same number of randomly selected tokens and refills by reward rank without excluding low-low tokens. Both preserve the retention ratio and OPD loss (Appendix~\ref{app:low-low-analysis}).

\paragraph{Results.}
(i) \emph{Low-low tokens carry disproportionate reward.} At 10\%, 15\%, and 20\% retention, they constitute 20.59\%, 13.49\%, and 10.26\% of selected tokens but contribute 49.80\%, 41.43\%, and 38.01\% of total absolute reward, respectively (Figure~\ref{fig:low-low-dynamics}). (ii) \emph{Targeted replacement accelerates learning and improves accuracy.} Low-low replacement yields earlier gains and accuracies of 38.4\%, 37.6\%, and 40.8\%, exceeding Reward-Top by 12.8, 7.2, and 6.4 percentage points, respectively (Figure~\ref{fig:low-low-replacement}). It also consistently outperforms random replacement, supporting targeted low-low removal. Thus, \textbf{low-low supervision can hinder learning despite large rewards}, indicating that \emph{token selection should consider both relative disagreement and absolute probability scale.}

\section{Method}
\label{sec:method}

Reward magnitude helps identify valuable tokens, yet low-low tokens contribute disproportionate absolute reward and can hinder learning. We therefore propose \textbf{DIAL-OPD}, a token-selection method that \emph{preserves the relative-disagreement signal captured by reward magnitude while incorporating absolute probability scale to downweight low-low tokens}.

\subsection{Bridging Log-Probability and Probability Spaces}
\label{sec:method-score}

Let $p_t=\pi_\theta(y_t\mid c_t)$ and $q_t=\pi_T(y_t\mid c_t)$ denote student and teacher probabilities for a sampled token. To make log-ratio-based selection probability-aware, we use the logarithmic mean~\citep{bhatia2008logmean}, which averages probabilities along a uniform interpolation between $\log p_t$ and $\log q_t$:
\begin{equation}
L(p_t,q_t)=\int_0^1p_t^{1-u}q_t^u\,du,
\label{eq:logarithmic-mean}
\end{equation}
where $u\in[0,1]$ parameterizes the interpolation and $p_t^{1-u}q_t^u$ is the corresponding probability. Since these probabilities lie between $p_t$ and $q_t$, \emph{low probabilities under both models yield a small logarithmic mean}. \method{} incorporates this probability scale by weighting the OPD reward magnitude with $L(p_t,q_t)^\beta$, where $\beta$ controls the weighting strength:
\begin{equation}
\boxed{\ s_{\beta,t}=L(p_t,q_t)^\beta\bigl|r_t^{\mathrm{OPD}}\bigr|\ },
\qquad \beta\geq0.
\label{eq:dial-score}
\end{equation}

For $\beta>0$, the small logarithmic mean of low-low tokens reduces their selection scores. Substituting $L(p_t,q_t)=(q_t-p_t)/(\log q_t-\log p_t)$ for $p_t\ne q_t$ connects relative and absolute disagreement:
\begin{equation}
s_{\beta,t}
=\left(\frac{|q_t-p_t|}{|r_t^{\mathrm{OPD}}|}\right)^\beta
\bigl|r_t^{\mathrm{OPD}}\bigr|
=\bigl|r_t^{\mathrm{OPD}}\bigr|^{1-\beta}|q_t-p_t|^\beta.
\label{eq:dial-endpoints}
\end{equation}
At $\beta=0$ and $\beta=1$, selection uses the OPD reward magnitude and absolute probability gap, respectively. Intermediate values geometrically interpolate between them, while $\beta>1$ strengthens probability weighting. \textbf{DIAL-OPD thus provides a continuous dial between relative and absolute disagreement for token selection.}

\subsection{The Role of \texorpdfstring{$\beta$}{beta} in DIAL-OPD}
\label{sec:method-properties}
\begin{figure}[t]
\centering
\includegraphics[width=\linewidth]{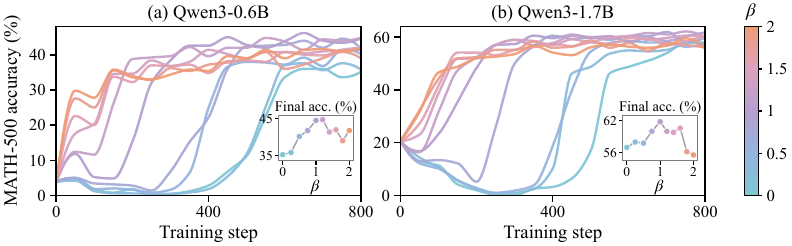}\par
\vspace{-10pt}
\caption{\small Learning dynamics of DIAL-OPD across $\beta$ for Qwen3-0.6B and Qwen3-1.7B students, using a Qwen3-4B teacher and 40\% token retention. Insets show final accuracy as a function of $\beta$.}
\label{fig:beta-curves}
\end{figure}

We examine how $\beta$ affects token selection when (i) student and teacher probabilities are both low, or (ii) one is low and the other is moderate or high.
\begin{proposition}[Controllable Suppression of Low-Low Disagreement]
\label{prop:low-low-scaling}
For $p,q\in(0,1]$ and $\beta\geq0$, scaling both probabilities by $\varepsilon\in(0,1]$ changes the selection score according to
\begin{equation}
s_\beta(\varepsilon p,\varepsilon q)=\varepsilon^\beta s_\beta(p,q).
\label{eq:probability-scaling}
\end{equation}
\end{proposition}
For $\varepsilon<1$ and $p\ne q$, the score remains unchanged at $\beta=0$, while larger $\beta>0$ increasingly suppresses low-low disagreement. We next examine when only one model assigns low probability.
\begin{proposition}[Sensitivity to One-Sided Low Probabilities]
\label{prop:one-sided-low-probabilities}
For fixed $a\in(0,1]$ and $\beta\geq0$,
\begin{equation}
\lim_{b\to0^+}s_\beta(a,b)
=\lim_{b\to0^+}s_\beta(b,a)
=\begin{cases}
+\infty, & 0\leq\beta<1,\\[-2pt]
a, & \beta=1,\\[-2pt]
0, & \beta>1.
\end{cases}
\label{eq:vanishing-student}
\end{equation}
\end{proposition}

Proofs are given in Appendices~\ref{app:proposition-one-proof} and~\ref{app:proposition-two-proof}. Together, these propositions reveal a trade-off: increasing $\beta$ strengthens low-low suppression, but $\beta>1$ also drives scores to zero when only one probability vanishes. Only $\beta=1$ preserves a finite, nonzero limit in this case.

\subsection{DIAL-OPD}
\label{sec:method-training}

Let $\mathcal V_i$ denote the candidate token positions in student-generated response $i$, with $|\mathcal V_i|>0$. At retention ratio $\rho\in(0,1]$, \method{} selects $K_i$ tokens maximizing their total score:
\begin{equation}
K_i=\max\{1,\operatorname{round}(\rho|\mathcal V_i|)\},
\quad
\mathcal S_i^\star\in
\underset{\substack{\mathcal S\subseteq\mathcal V_i\\|\mathcal S|=K_i}}{\arg\max}
\sum_{t\in\mathcal S}s_{\beta,i,t},
\quad
m_{i,t}=\mathbb I\{t\in\mathcal S_i^\star\}.
\label{eq:dial-selection}
\end{equation}
Using the binary selection mask $m_{i,t}$, \method{} applies the original signed OPD reward (Section~\ref{sec:prelim-reward}) only to selected tokens. For a training batch $\mathcal B$, it minimizes the masked objective:
\begin{equation}
\mathcal L_{\mathrm{DIAL}}(\theta)
=-\frac{1}{M_{\mathcal B}}
\sum_{i\in\mathcal B}\sum_{t\in\mathcal V_i}
m_{i,t}\operatorname{sg}[r^{\mathrm{OPD}}_{i,t}]
\log\pi_\theta(y_{i,t}\mid c_{i,t}),
\qquad
M_{\mathcal B}=\sum_{i\in\mathcal B}\sum_{t\in\mathcal V_i}m_{i,t}.
\label{eq:dial-loss}
\end{equation}
Here, $\operatorname{sg}$ denotes stop-gradient, with the mask fixed during differentiation. Unselected tokens remain in context but are excluded from the loss. \method{} recomputes scores and masks at each training step using fresh rollouts from the current student policy.

\section{Experimental Setup}
\label{sec:experimental-setup}
\begin{table}[t]
\centering
\caption{Results with a Qwen3-4B teacher on seven benchmarks. \textbf{Bold} and \underline{underlined} mark the best and second-best accuracy per student and retention ratio, excluding full-token references.}
\label{tab:main-t4b}
\begingroup
\fontsize{6.7}{8.4}\selectfont
\setlength{\tabcolsep}{0.50pt}
\renewcommand{\arraystretch}{0.94}
\setlength{\aboverulesep}{1.0pt}
\setlength{\belowrulesep}{1.0pt}
\sbox{\resultstablebox}{%
\begin{tabular}{@{}c@{\hspace{2.5pt}}l@{\hspace{2.5pt}}*{7}{rr>{\color{black!64}}r@{\hspace{4pt}}}r@{}}
\toprule
Keep & Method & \multicolumn{3}{c}{GSM8K} & \multicolumn{3}{c}{MATH500} & \multicolumn{3}{c}{AMC23} & \multicolumn{3}{c}{AIME24} & \multicolumn{3}{c}{AIME25} & \multicolumn{3}{c}{Minerva} & \multicolumn{3}{c}{Olympiad} & Mean \\
\cmidrule(lr){3-5} \cmidrule(lr){6-8} \cmidrule(lr){9-11} \cmidrule(lr){12-14} \cmidrule(lr){15-17} \cmidrule(lr){18-20} \cmidrule(lr){21-23}
 & & Avg & Pass & Len & Avg & Pass & Len & Avg & Pass & Len & Avg & Pass & Len & Avg & Pass & Len & Avg & Pass & Len & Avg & Pass & Len & Avg \\
\midrule
\rowcolor{studentband}
\multicolumn{24}{c}{\rule{0pt}{8.5pt}\textbf{Student: Qwen3-0.6B}\rule[-2pt]{0pt}{0pt}} \\
\midrule
 & Vanilla OPD & 58.5 & 79.7 & 506 & 35.9 & 52.6 & 1687 & 17.3 & 50.0 & 5059 & 1.0 & 6.7 & 7266 & 1.2 & 10.0 & 7369 & 9.2 & 22.8 & 1436 & 12.8 & 24.5 & 5412 & 19.4 \\
100\% & PowerOPD & 68.5 & 84.9 & 428 & 43.5 & 60.0 & 1511 & 23.6 & 62.5 & 4338 & 1.7 & 3.3 & 6535 & 1.0 & 10.0 & 6816 & 10.9 & 22.8 & 1229 & 17.5 & 30.3 & 4531 & 23.8 \\
 & ExOPD & 54.8 & 78.8 & 533 & 34.4 & 52.6 & 1876 & 13.1 & 45.0 & 5512 & 1.7 & 6.7 & 7416 & 0.6 & 10.0 & 7704 & 8.2 & 20.2 & 1702 & 12.2 & 24.5 & 5565 & 17.9 \\
\midrule
 & Random selection & 38.8 & 70.7 & 1880 & 12.0 & 32.0 & 3196 & 5.6 & 32.5 & 6893 & 0.0 & 0.0 & 7787 & 0.4 & \underline{6.7} & 7555 & 4.0 & 15.4 & 2996 & 4.8 & 14.4 & 7057 & 9.4 \\
 & TIP & 54.7 & 76.4 & 610 & 29.9 & 48.0 & 2058 & 10.8 & \textbf{55.0} & 5663 & 0.8 & \underline{3.3} & 7611 & 0.2 & 3.3 & 7801 & 6.8 & 19.1 & 1856 & 10.9 & 22.8 & 5856 & 16.3 \\
 & SA-OPD & 56.6 & 80.0 & 533 & \underline{35.5} & \underline{52.4} & 1731 & 15.6 & \underline{52.5} & 5142 & \underline{1.2} & \textbf{6.7} & 6979 & \underline{1.0} & \textbf{10.0} & 7163 & \textbf{9.3} & \underline{22.8} & 1429 & \underline{13.8} & \underline{26.6} & 5296 & 19.0 \\
 & TA-OPD & 7.6 & 25.8 & 3160 & 5.0 & 17.4 & 3386 & 2.2 & 30.0 & 6711 & 0.0 & 0.0 & 7097 & 0.2 & 3.3 & 7209 & 1.7 & 9.9 & 3218 & 2.1 & 7.4 & 6876 & 2.7 \\
20\% & Prefix OPD & 45.6 & 75.5 & 1087 & 23.8 & 47.8 & 2307 & 11.4 & 47.5 & 5736 & 0.4 & \underline{3.3} & 7041 & 0.8 & \underline{6.7} & 7290 & 5.8 & 19.5 & 2133 & 9.4 & 23.6 & 5773 & 13.9 \\
 & CC-OPD & 50.2 & 66.9 & 2615 & 21.7 & 35.0 & 3523 & 6.4 & 27.5 & 7934 & 0.0 & 0.0 & 8192 & \underline{1.0} & 3.3 & 8192 & 4.4 & 14.0 & 3775 & 6.2 & 13.8 & 7883 & 12.8 \\
\cmidrule(lr){2-24}
\rowcolor{methodblue}
\cellcolor{white} & \textbf{DIAL-OPD ($\beta{=}1$)} & \textbf{67.0} & \underline{81.3} & 1072 & 34.2 & 47.2 & 2609 & \underline{15.9} & 45.0 & 6554 & 0.4 & \textbf{6.7} & 7985 & \textbf{1.5} & \textbf{10.0} & 8016 & \underline{7.6} & 16.9 & 2617 & 13.1 & 24.2 & 6658 & \underline{19.9} \\
\rowcolor{methodblue}
\cellcolor{white} & \textbf{DIAL-OPD ($\beta{=}1.2$)} & \underline{63.4} & \textbf{82.6} & 553 & \textbf{39.7} & \textbf{54.6} & 1793 & \textbf{17.3} & \underline{52.5} & 4910 & \textbf{1.9} & \underline{3.3} & 6874 & 0.8 & \textbf{10.0} & 7239 & \textbf{9.3} & \textbf{23.9} & 1522 & \textbf{13.9} & \textbf{27.4} & 4907 & \textbf{20.9} \\
\midrule
 & Random selection & 53.8 & 80.0 & 610 & 33.7 & 52.0 & 1997 & 14.1 & 52.5 & 5545 & 0.6 & 3.3 & 7376 & 0.6 & 6.7 & 7381 & 8.0 & \underline{22.1} & 1730 & 11.8 & 24.5 & 5724 & 17.5 \\
 & TIP & 56.6 & 79.2 & 576 & 35.1 & 51.4 & 1829 & 16.9 & \textbf{65.0} & 5321 & 1.0 & 3.3 & 7405 & 0.8 & 6.7 & 7666 & 8.6 & 21.7 & 1622 & 12.3 & 24.8 & 5520 & 18.8 \\
 & SA-OPD & 55.8 & 79.8 & 500 & 37.0 & 53.0 & 1749 & 14.7 & 55.0 & 5142 & 1.0 & 3.3 & 6997 & 0.6 & \underline{10.0} & 7447 & 8.5 & 21.0 & 1491 & 12.6 & 24.5 & 5491 & 18.6 \\
 & TA-OPD & 11.9 & 38.1 & 2574 & 7.8 & 24.6 & 2782 & 3.0 & 27.5 & 5864 & 0.2 & 3.3 & 6250 & 0.2 & 3.3 & 6177 & 2.6 & 15.1 & 2744 & 2.7 & 9.2 & 6042 & 4.1 \\
40\% & Prefix OPD & 50.3 & 78.0 & 754 & 32.2 & 50.2 & 2099 & 13.8 & 52.5 & 5715 & 0.4 & 3.3 & 7379 & 0.4 & 3.3 & 7596 & 8.7 & 21.3 & 1863 & 10.2 & 21.8 & 5964 & 16.6 \\
 & CC-OPD & \underline{67.5} & 82.7 & 1120 & 39.9 & 53.6 & 2312 & 22.7 & 55.0 & 6381 & 1.7 & \underline{10.0} & 7916 & \textbf{2.3} & \textbf{13.3} & 7931 & 8.9 & \underline{22.1} & 2302 & 16.1 & 27.0 & 6225 & 22.7 \\
\cmidrule(lr){2-24}
\rowcolor{methodblue}
\cellcolor{white} & \textbf{DIAL-OPD ($\beta{=}1$)} & \textbf{69.1} & \textbf{84.7} & 578 & \textbf{44.9} & \textbf{61.6} & 1714 & \textbf{25.5} & 50.0 & 4774 & \textbf{2.5} & \textbf{13.3} & 7001 & \underline{1.2} & 6.7 & 7179 & \underline{10.7} & \underline{22.1} & 1484 & \textbf{18.8} & \textbf{33.5} & 4705 & \textbf{24.7} \\
\rowcolor{methodblue}
\cellcolor{white} & \textbf{DIAL-OPD ($\beta{=}1.2$)} & 67.2 & \underline{83.5} & 439 & \underline{44.0} & \underline{59.0} & 1521 & \underline{25.2} & \underline{62.5} & 4520 & \underline{2.1} & \underline{10.0} & 6761 & 0.2 & 3.3 & 6884 & \textbf{10.9} & \textbf{25.7} & 1216 & \underline{17.1} & \underline{29.4} & 4604 & \underline{23.8} \\
\addlinespace[1pt]
\midrule[0.8pt]
\rowcolor{studentband}
\multicolumn{24}{c}{\rule{0pt}{8.5pt}\textbf{Student: Qwen3-1.7B}\rule[-2pt]{0pt}{0pt}} \\
\midrule
 & Vanilla OPD & 74.2 & 92.6 & 348 & 58.0 & 73.4 & 1150 & 33.8 & 72.5 & 3471 & 7.7 & 16.7 & 6150 & 7.3 & 26.7 & 5791 & 17.2 & 30.9 & 1018 & 26.6 & 41.2 & 3887 & 32.1 \\
100\% & PowerOPD & 81.5 & 92.6 & 340 & 59.6 & 74.2 & 1208 & 35.5 & 75.0 & 3453 & 7.9 & 20.0 & 5946 & 7.5 & 26.7 & 5668 & 18.4 & 32.0 & 992 & 28.5 & 43.3 & 3768 & 34.1 \\
 & ExOPD & 65.0 & 90.6 & 327 & 47.3 & 68.6 & 1265 & 25.9 & 65.0 & 3653 & 6.2 & 23.3 & 5951 & 4.8 & 23.3 & 5435 & 14.1 & 30.1 & 1047 & 22.3 & 39.0 & 3843 & 26.5 \\
\midrule
 & Random selection & 36.4 & 78.7 & 1173 & 15.8 & 40.6 & 2126 & 7.3 & 47.5 & 5092 & 1.0 & 13.3 & 5719 & 2.5 & 13.3 & 5482 & 4.9 & 23.2 & 2285 & 6.3 & 19.1 & 4847 & 10.6 \\
 & TIP & 64.3 & 89.8 & 346 & 47.4 & 70.0 & 1167 & 25.6 & \textbf{75.0} & 3561 & 6.0 & \underline{20.0} & 5818 & 3.8 & 16.7 & 5516 & 13.0 & 28.3 & 987 & 19.4 & 35.5 & 3725 & 25.7 \\
 & SA-OPD & 68.9 & 91.0 & 367 & 52.9 & \underline{70.4} & 1155 & 30.2 & \textbf{75.0} & 3529 & 5.6 & 16.7 & 5592 & \underline{4.8} & \textbf{26.7} & 5545 & 15.7 & 30.5 & 931 & 23.8 & 38.6 & 3617 & 28.8 \\
 & TA-OPD & 42.6 & 83.0 & 1472 & 35.1 & 67.6 & 1930 & 18.0 & 62.5 & 4506 & 5.4 & \underline{20.0} & 5738 & 4.0 & 16.7 & 5472 & 10.4 & 27.6 & 1854 & 16.9 & 34.9 & 4433 & 18.9 \\
20\% & Prefix OPD & 54.3 & 86.6 & 459 & 36.6 & 65.6 & 1218 & 21.6 & 62.5 & 3101 & 4.6 & \underline{20.0} & 5331 & 3.3 & 16.7 & 4745 & 10.9 & 29.0 & 1170 & 15.9 & 34.0 & 3100 & 21.0 \\
 & CC-OPD & 79.6 & \underline{92.8} & 662 & 48.4 & 62.2 & 1949 & 28.9 & 62.5 & 5556 & 4.4 & \underline{20.0} & 7717 & \underline{4.8} & 16.7 & 7665 & 15.7 & \underline{31.6} & 2030 & 20.3 & 32.2 & 5736 & 28.9 \\
\cmidrule(lr){2-24}
\rowcolor{methodblue}
\cellcolor{white} & \textbf{DIAL-OPD ($\beta{=}1$)} & \textbf{82.3} & 91.8 & 404 & \textbf{57.8} & 70.2 & 1445 & \textbf{37.8} & \textbf{75.0} & 4276 & \textbf{8.8} & \textbf{26.7} & 6287 & \underline{4.8} & 16.7 & 6170 & \textbf{17.9} & \textbf{32.7} & 1216 & \textbf{29.5} & \textbf{43.2} & 4222 & \textbf{34.1} \\
\rowcolor{methodblue}
\cellcolor{white} & \textbf{DIAL-OPD ($\beta{=}1.2$)} & \underline{81.7} & \textbf{93.5} & 342 & \underline{57.7} & \textbf{73.0} & 1288 & \underline{35.9} & \underline{72.5} & 3640 & \underline{7.5} & \underline{20.0} & 6040 & \textbf{6.9} & \underline{23.3} & 5730 & \underline{17.0} & 30.9 & 1045 & \underline{26.5} & \underline{42.6} & 3911 & \underline{33.3} \\
\midrule
 & Random selection & 63.8 & 90.8 & 349 & 50.1 & 72.2 & 1076 & 28.3 & 70.0 & 3229 & 6.2 & \textbf{26.7} & 5177 & 5.6 & 13.3 & 4830 & 15.4 & 30.5 & 947 & 23.0 & 40.4 & 3382 & 27.5 \\
 & TIP & 70.8 & 91.4 & 333 & 50.7 & 70.4 & 1151 & 29.2 & \underline{72.5} & 3264 & 7.9 & \underline{23.3} & 5789 & 6.5 & \underline{20.0} & 5571 & 14.2 & 28.3 & 933 & 23.7 & 38.7 & 3765 & 29.0 \\
 & SA-OPD & 68.9 & 91.6 & 343 & 54.3 & 72.2 & 1104 & 30.5 & \underline{72.5} & 3605 & 7.7 & 20.0 & 5885 & \textbf{7.7} & 16.7 & 5210 & 16.3 & \underline{32.0} & 913 & 25.1 & 41.2 & 3634 & 30.1 \\
 & TA-OPD & 9.2 & 31.2 & 2890 & 6.7 & 23.6 & 3040 & 2.7 & 27.5 & 6197 & 0.8 & 10.0 & 6496 & 0.2 & 3.3 & 6174 & 2.8 & 15.8 & 2957 & 2.7 & 9.9 & 6384 & 3.6 \\
40\% & Prefix OPD & 57.7 & 88.5 & 435 & 40.0 & 67.6 & 1364 & 23.4 & 65.0 & 3994 & 5.4 & 20.0 & 6371 & 4.2 & 16.7 & 5823 & 12.4 & 27.2 & 1256 & 19.5 & 36.5 & 3980 & 23.2 \\
 & CC-OPD & 81.6 & \underline{92.8} & 376 & 59.2 & \underline{73.2} & 1337 & 34.2 & 70.0 & 4265 & 8.1 & \underline{23.3} & 6608 & \underline{7.5} & \textbf{26.7} & 6483 & 18.2 & 29.4 & 1077 & \underline{28.2} & \underline{42.1} & 4339 & 33.9 \\
\cmidrule(lr){2-24}
\rowcolor{methodblue}
\cellcolor{white} & \textbf{DIAL-OPD ($\beta{=}1$)} & \textbf{83.2} & \underline{92.8} & 344 & \textbf{61.5} & 73.0 & 1223 & \textbf{38.8} & \textbf{75.0} & 3457 & \underline{9.8} & \textbf{26.7} & 5992 & 5.2 & 16.7 & 5951 & \textbf{19.4} & 30.9 & 992 & \textbf{29.7} & \textbf{44.5} & 3789 & \textbf{35.4} \\
\rowcolor{methodblue}
\cellcolor{white} & \textbf{DIAL-OPD ($\beta{=}1.2$)} & \underline{82.5} & \textbf{93.0} & 330 & \underline{61.0} & \textbf{74.4} & 1199 & \underline{36.6} & \textbf{75.0} & 3679 & \textbf{10.0} & \textbf{26.7} & 6191 & 7.3 & \textbf{26.7} & 5934 & \underline{19.1} & \textbf{33.1} & 977 & 27.8 & 40.9 & 3887 & \underline{34.9} \\
\bottomrule
\end{tabular}}
\typeout{STACKED-WIDTH=\the\wd\resultstablebox; LIMIT=\the\linewidth}
\typeout{STACKED-HEIGHT=\the\dimexpr\ht\resultstablebox+\dp\resultstablebox\relax; TEXTHEIGHT=\the\textheight}
\usebox{\resultstablebox}
\endgroup
\end{table}

\begin{table}[t]
\centering
\caption{Results with a Qwen3-8B teacher on seven benchmarks. \textbf{Bold} and \underline{underlined} mark the best and second-best accuracy per student and retention ratio, excluding full-token references.}
\label{tab:main-t8b}
\begingroup
\fontsize{6.7}{8.4}\selectfont
\setlength{\tabcolsep}{0.50pt}
\renewcommand{\arraystretch}{0.94}
\setlength{\aboverulesep}{1.0pt}
\setlength{\belowrulesep}{1.0pt}
\sbox{\resultstablebox}{%
\begin{tabular}{@{}c@{\hspace{2.5pt}}l@{\hspace{2.5pt}}*{7}{rr>{\color{black!64}}r@{\hspace{4pt}}}r@{}}
\toprule
Keep & Method & \multicolumn{3}{c}{GSM8K} & \multicolumn{3}{c}{MATH500} & \multicolumn{3}{c}{AMC23} & \multicolumn{3}{c}{AIME24} & \multicolumn{3}{c}{AIME25} & \multicolumn{3}{c}{Minerva} & \multicolumn{3}{c}{Olympiad} & Mean \\
\cmidrule(lr){3-5} \cmidrule(lr){6-8} \cmidrule(lr){9-11} \cmidrule(lr){12-14} \cmidrule(lr){15-17} \cmidrule(lr){18-20} \cmidrule(lr){21-23}
 & & Avg & Pass & Len & Avg & Pass & Len & Avg & Pass & Len & Avg & Pass & Len & Avg & Pass & Len & Avg & Pass & Len & Avg & Pass & Len & Avg \\
\midrule
\rowcolor{studentband}
\multicolumn{24}{c}{\rule{0pt}{8.5pt}\textbf{Student: Qwen3-0.6B}\rule[-2pt]{0pt}{0pt}} \\
\midrule
 & Vanilla OPD & 58.4 & 81.1 & 483 & 39.2 & 56.8 & 1573 & 16.2 & 57.5 & 4488 & 1.7 & 6.7 & 6651 & 1.0 & 6.7 & 7033 & 9.1 & 21.0 & 1327 & 14.1 & 26.9 & 4864 & 20.0 \\
100\% & PowerOPD & 67.2 & 83.7 & 404 & 43.8 & 61.2 & 1513 & 23.8 & 57.5 & 4249 & 2.3 & 10.0 & 6417 & 1.5 & 10.0 & 6577 & 10.6 & 21.7 & 1163 & 17.0 & 30.9 & 4415 & 23.7 \\
 & ExOPD & 57.0 & 79.8 & 442 & 38.1 & 57.0 & 1568 & 17.7 & 52.5 & 4496 & 2.3 & 13.3 & 6715 & 0.8 & 6.7 & 7153 & 8.7 & 22.8 & 1303 & 13.5 & 26.1 & 5076 & 19.7 \\
\midrule
 & Random selection & 38.2 & 70.9 & 1314 & 9.9 & 27.0 & 2879 & 6.2 & 45.0 & 6345 & 0.0 & 0.0 & 7430 & 1.0 & \underline{10.0} & 7201 & 3.5 & 15.8 & 2420 & 3.7 & 11.3 & 6611 & 9.0 \\
 & TIP & 56.8 & 81.0 & 515 & 34.4 & 51.8 & 1706 & 13.1 & 42.5 & 5191 & \underline{1.5} & \underline{3.3} & 7147 & 0.8 & 6.7 & 7231 & 8.5 & 21.7 & 1526 & 12.8 & \underline{26.3} & 5333 & 18.3 \\
 & SA-OPD & 59.2 & 79.8 & 463 & 40.0 & 57.0 & 1596 & 19.5 & 55.0 & 4603 & \underline{1.5} & \underline{3.3} & 6502 & \underline{1.7} & \textbf{13.3} & 7003 & \underline{9.4} & \underline{22.8} & 1301 & \underline{13.7} & 25.7 & 4911 & 20.7 \\
 & TA-OPD & 8.6 & 28.1 & 3109 & 6.0 & 19.8 & 3286 & 2.8 & 30.0 & 6722 & 0.4 & \underline{3.3} & 6910 & 0.4 & 3.3 & 7076 & 1.8 & 10.7 & 3148 & 2.9 & 9.9 & 6795 & 3.3 \\
20\% & Prefix OPD & 49.7 & 77.0 & 722 & 30.6 & 51.2 & 1889 & 13.4 & 50.0 & 5274 & 0.8 & \underline{3.3} & 6911 & 1.2 & \textbf{13.3} & 7144 & 8.2 & 22.1 & 1597 & 10.2 & 23.3 & 5256 & 16.3 \\
 & CC-OPD & 54.3 & 73.1 & 2035 & 26.3 & 39.4 & 3219 & 9.7 & 32.5 & 7672 & 0.0 & 0.0 & 8179 & 1.0 & 3.3 & 8188 & 5.1 & 15.8 & 3464 & 7.1 & 14.8 & 7654 & 14.8 \\
\cmidrule(lr){2-24}
\rowcolor{methodblue}
\cellcolor{white} & \textbf{DIAL-OPD ($\beta{=}1$)} & \underline{63.8} & \underline{83.2} & 465 & \textbf{42.3} & \underline{59.6} & 1633 & \underline{21.2} & \textbf{62.5} & 4372 & \textbf{1.7} & \textbf{10.0} & 6566 & \textbf{2.1} & \underline{10.0} & 6583 & \textbf{9.7} & \textbf{23.2} & 1265 & \textbf{16.0} & \textbf{29.1} & 4494 & \textbf{22.4} \\
\rowcolor{methodblue}
\cellcolor{white} & \textbf{DIAL-OPD ($\beta{=}1.2$)} & \textbf{63.9} & \textbf{84.1} & 469 & \underline{40.5} & \textbf{60.0} & 1701 & \textbf{21.6} & \underline{57.5} & 4556 & 0.8 & \underline{3.3} & 6544 & 1.0 & \underline{10.0} & 6786 & 8.7 & \textbf{23.2} & 1351 & 13.6 & 25.7 & 4567 & \underline{21.4} \\
\midrule
 & Random selection & 55.9 & 79.7 & 627 & 36.4 & 54.0 & 1715 & 15.8 & 47.5 & 4963 & 1.5 & 3.3 & 6809 & 1.5 & \underline{10.0} & 7066 & 8.9 & \underline{22.8} & 1449 & 13.7 & 26.6 & 4988 & 19.1 \\
 & TIP & 59.2 & 81.6 & 468 & 39.6 & 57.8 & 1587 & 18.4 & 57.5 & 4665 & 1.9 & \underline{10.0} & 7000 & 0.8 & \underline{10.0} & 7065 & 8.6 & \underline{22.8} & 1344 & 14.5 & 26.4 & 5132 & 20.4 \\
 & SA-OPD & 58.8 & 80.7 & 459 & 39.7 & 56.4 & 1529 & 19.2 & 60.0 & 4517 & 2.3 & \underline{10.0} & 6681 & 0.2 & 3.3 & 7151 & 9.4 & 19.9 & 1261 & 15.6 & 26.7 & 4739 & 20.7 \\
 & TA-OPD & 17.3 & 48.7 & 2615 & 12.1 & 34.0 & 2881 & 7.5 & 47.5 & 6181 & 0.8 & 6.7 & 6850 & 0.2 & 3.3 & 6944 & 3.8 & 17.6 & 2715 & 5.0 & 16.2 & 6151 & 6.7 \\
40\% & Prefix OPD & 54.5 & 79.8 & 627 & 37.0 & 56.6 & 1868 & 16.4 & 57.5 & 5483 & 0.6 & 3.3 & 7401 & 0.6 & 6.7 & 7555 & 9.2 & \textbf{23.9} & 1561 & 12.1 & 24.0 & 5597 & 18.6 \\
 & CC-OPD & \textbf{69.2} & \textbf{85.1} & 942 & 40.8 & 53.8 & 2163 & 23.6 & 52.5 & 6103 & \underline{2.9} & 6.7 & 7683 & \underline{1.7} & \underline{10.0} & 7972 & 9.4 & 21.3 & 2089 & 15.7 & 27.4 & 6170 & 23.3 \\
\cmidrule(lr){2-24}
\rowcolor{methodblue}
\cellcolor{white} & \textbf{DIAL-OPD ($\beta{=}1$)} & \underline{66.6} & \underline{84.6} & 436 & \textbf{45.6} & \textbf{61.4} & 1482 & \textbf{27.3} & \underline{62.5} & 3872 & \textbf{3.1} & \textbf{13.3} & 6432 & 1.2 & 6.7 & 6589 & \textbf{10.6} & 22.1 & 1197 & \underline{17.7} & \underline{29.7} & 4392 & \textbf{24.6} \\
\rowcolor{methodblue}
\cellcolor{white} & \textbf{DIAL-OPD ($\beta{=}1.2$)} & 65.6 & 82.7 & 438 & \underline{43.2} & \underline{60.4} & 1496 & \underline{26.7} & \textbf{65.0} & 4165 & 1.5 & 3.3 & 6162 & \textbf{1.9} & \textbf{13.3} & 6526 & \underline{10.4} & \underline{22.8} & 1216 & \textbf{18.1} & \textbf{31.0} & 4376 & \underline{23.9} \\
\addlinespace[1pt]
\midrule[0.8pt]
\rowcolor{studentband}
\multicolumn{24}{c}{\rule{0pt}{8.5pt}\textbf{Student: Qwen3-1.7B}\rule[-2pt]{0pt}{0pt}} \\
\midrule
 & Vanilla OPD & 78.7 & 92.5 & 348 & 59.5 & 72.8 & 1201 & 34.7 & 77.5 & 3457 & 7.5 & 20.0 & 6135 & 4.2 & 13.3 & 5501 & 17.9 & 30.5 & 983 & 27.7 & 42.1 & 3718 & 32.9 \\
100\% & PowerOPD & 80.1 & 92.4 & 327 & 58.3 & 71.2 & 1160 & 35.6 & 72.5 & 3343 & 9.2 & 30.0 & 5593 & 6.7 & 26.7 & 5118 & 16.8 & 31.2 & 925 & 27.6 & 40.5 & 3531 & 33.5 \\
 & ExOPD & 80.3 & 92.3 & 337 & 58.7 & 73.0 & 1191 & 35.2 & 70.0 & 3413 & 9.2 & 26.7 & 5640 & 6.2 & 20.0 & 5323 & 18.0 & 28.7 & 977 & 27.7 & 41.8 & 3721 & 33.6 \\
\midrule
 & Random selection & 66.8 & 91.0 & 541 & 49.0 & 70.0 & 1290 & 26.4 & 65.0 & 3323 & 6.0 & 16.7 & 5586 & 5.2 & \textbf{23.3} & 5176 & 15.5 & 30.9 & 1164 & 22.3 & 37.7 & 3572 & 27.3 \\
 & TIP & 74.9 & 91.8 & 354 & 54.6 & 71.4 & 1167 & 28.4 & \textbf{75.0} & 3794 & \textbf{10.0} & \textbf{26.7} & 5756 & 6.2 & 16.7 & 5284 & 15.7 & 29.8 & 942 & 24.1 & 38.7 & 3785 & 30.6 \\
 & SA-OPD & 75.9 & 91.4 & 331 & 58.6 & \underline{73.8} & 1093 & 31.9 & \textbf{75.0} & 3240 & 6.2 & 20.0 & 5838 & \underline{6.7} & \textbf{23.3} & 5009 & 17.1 & \textbf{32.0} & 905 & 25.7 & 40.7 & 3540 & 31.7 \\
 & TA-OPD & \underline{80.5} & \textbf{93.3} & 337 & 58.4 & \textbf{74.4} & 1066 & \underline{35.5} & \textbf{75.0} & 2977 & \underline{9.0} & \underline{23.3} & 5310 & 6.5 & \underline{20.0} & 4810 & \underline{17.9} & \textbf{32.0} & 895 & 26.1 & 40.8 & 3163 & \underline{33.4} \\
20\% & Prefix OPD & 76.1 & 91.2 & 390 & 52.2 & 66.0 & 1181 & 31.4 & 67.5 & 3355 & 5.8 & \underline{23.3} & 5872 & 5.6 & \textbf{23.3} & 5303 & 17.1 & \underline{31.6} & 1017 & 21.8 & 34.3 & 3528 & 30.0 \\
 & CC-OPD & 76.2 & 91.8 & 567 & 49.8 & 67.2 & 1770 & 29.5 & 60.0 & 5146 & 5.0 & 20.0 & 7371 & 5.4 & 16.7 & 7150 & 14.4 & 30.9 & 1645 & 21.8 & 34.9 & 5313 & 28.9 \\
\cmidrule(lr){2-24}
\rowcolor{methodblue}
\cellcolor{white} & \textbf{DIAL-OPD ($\beta{=}1$)} & \textbf{80.6} & \underline{92.9} & 333 & \textbf{60.2} & \textbf{74.4} & 1199 & \textbf{35.8} & 70.0 & 3404 & 7.9 & 16.7 & 5791 & 5.4 & 16.7 & 5339 & \textbf{18.3} & 31.2 & 967 & \textbf{28.2} & \textbf{42.3} & 3570 & \textbf{33.8} \\
\rowcolor{methodblue}
\cellcolor{white} & \textbf{DIAL-OPD ($\beta{=}1.2$)} & 77.6 & 91.5 & 336 & \underline{59.1} & 73.6 & 1213 & 34.5 & \underline{72.5} & 3478 & 7.5 & \textbf{26.7} & 5793 & \textbf{6.9} & 16.7 & 5678 & 17.3 & \underline{31.6} & 992 & \underline{26.4} & \underline{41.8} & 3775 & 32.8 \\
\midrule
 & Random selection & 75.5 & 91.6 & 362 & 55.5 & 70.8 & 1141 & 30.8 & 67.5 & 3025 & \textbf{9.2} & \textbf{26.7} & 5670 & 4.4 & \textbf{26.7} & 5339 & 16.5 & 30.1 & 983 & 27.0 & 41.8 & 3518 & 31.3 \\
 & TIP & 75.7 & 91.1 & 346 & 58.1 & \textbf{74.2} & 1127 & 35.0 & 75.0 & 3422 & 7.9 & \textbf{26.7} & 5727 & 4.6 & 16.7 & 5441 & 17.6 & \textbf{32.7} & 961 & 26.0 & 41.8 & 3674 & 32.1 \\
 & SA-OPD & 76.7 & 91.1 & 330 & 59.0 & 72.6 & 1153 & 33.0 & \underline{80.0} & 3225 & 7.1 & 20.0 & 5637 & 5.4 & 20.0 & 4997 & 17.9 & 30.9 & 887 & 26.5 & 41.2 & 3584 & 32.2 \\
 & TA-OPD & 79.1 & 92.3 & 323 & 57.9 & 71.6 & 1113 & 34.2 & 72.5 & 3123 & \underline{8.8} & 16.7 & 5622 & \textbf{6.9} & 16.7 & 4898 & \textbf{18.2} & \underline{32.0} & 867 & 27.5 & 41.1 & 3460 & 33.2 \\
40\% & Prefix OPD & 76.9 & 92.2 & 359 & 54.3 & 67.4 & 1199 & 32.0 & 75.0 & 3696 & 7.9 & 20.0 & 6420 & 4.8 & \underline{23.3} & 5363 & 16.8 & 30.1 & 1011 & 22.1 & 34.9 & 3839 & 30.7 \\
 & CC-OPD & 79.9 & 92.5 & 346 & \underline{59.2} & 73.2 & 1245 & 35.8 & 72.5 & 3800 & 7.9 & \underline{23.3} & 6323 & 5.6 & \underline{23.3} & 6121 & 17.6 & 30.5 & 982 & 27.4 & 40.4 & 4047 & 33.3 \\
\cmidrule(lr){2-24}
\rowcolor{methodblue}
\cellcolor{white} & \textbf{DIAL-OPD ($\beta{=}1$)} & \textbf{82.4} & \textbf{93.5} & 323 & \textbf{59.7} & \underline{73.8} & 1173 & \textbf{38.0} & \textbf{82.5} & 3098 & \underline{8.8} & 16.7 & 5764 & \underline{6.5} & \underline{23.3} & 5425 & \underline{18.0} & 31.6 & 913 & \textbf{28.9} & \textbf{43.5} & 3533 & \textbf{34.6} \\
\rowcolor{methodblue}
\cellcolor{white} & \textbf{DIAL-OPD ($\beta{=}1.2$)} & \underline{82.1} & \underline{92.9} & 331 & \textbf{59.7} & 72.4 & 1156 & \underline{36.4} & 77.5 & 3134 & \textbf{9.2} & 20.0 & 5823 & 6.0 & \textbf{26.7} & 5306 & 17.1 & 28.7 & 898 & \underline{27.8} & \underline{42.4} & 3536 & \underline{34.1} \\
\bottomrule
\end{tabular}}
\typeout{STACKED-WIDTH=\the\wd\resultstablebox; LIMIT=\the\linewidth}
\typeout{STACKED-HEIGHT=\the\dimexpr\ht\resultstablebox+\dp\resultstablebox\relax; TEXTHEIGHT=\the\textheight}
\usebox{\resultstablebox}
\endgroup
\end{table}

\paragraph{Models and training.}
We train Qwen3 students (0.6B/1.7B) with 4B/8B teachers~\citep{qwen2025qwen3} on DeepScaleR~\citep{deepscaler2025} for 800 steps (details in Appendix~\ref{app:experimental-details}).

\paragraph{Evaluation.}
We evaluate on GSM8K~\citep{cobbe2021gsm8k}, MATH-500~\citep{lightman2023verify}, AMC23, AIME24, AIME25~\citep{maaCompetitions}, Minerva Math~\citep{lewkowycz2022minerva}, and OlympiadBench~\citep{he2024olympiadbench}.

\paragraph{Baselines.}
\textbf{(i)} \emph{Reference baselines} include full-token Vanilla OPD, uniform Random selection, PowerOPD with bounded power-transformed rewards~\citep{zhao2026poweropd}, and ExOPD with extrapolated teacher--reference rewards~\citep{yang2026exopd}. \textbf{(ii)} \emph{Token-selection baselines} include TIP, combining student entropy and teacher--student divergence~\citep{xu2026tip}; SA-OPD, using input dependence and distillation divergence~\citep{jiang2026saopd}; TA-OPD, combining disagreement and teacher support~\citep{wang2026taopd}; and Prefix OPD, retaining response prefixes~\citep{zhang2026prefixopd}. We also include Capability-Contrast OPD (CC-OPD), adapting Direct-OPD's policy contrast~\citep{feng2026directopd} through absolute teacher--reference probability gaps. Token-selection methods are compared at matched retention ratios; implementation details appear in Appendix~\ref{app:baseline-implementations}.

\section{Experiments}
\label{sec:experiments}

\subsection{The Effect of \texorpdfstring{$\beta$}{beta} on Learning}
\label{sec:beta-ablation}

DIAL-OPD uses probability scale to downweight low-low tokens, with $\beta$ controlling the weighting strength. Increasing $\beta$ suppresses low-low disagreement more strongly, but may also reduce the priority of useful disagreements supported by one model. We therefore examine how this trade-off affects learning dynamics and final accuracy across different student model sizes.

\paragraph{Experimental setup.}
We fix the retention ratio at 40\% and sweep $\beta$ over
$\{0,\allowbreak\,0.25,\allowbreak\,0.5,\allowbreak\,0.75,\allowbreak\,1,\allowbreak\,1.2,\allowbreak\,1.4,\allowbreak\,1.6,\allowbreak\,1.8,\allowbreak\,2\}$.
We use Qwen3-4B as the teacher and Qwen3-0.6B and Qwen3-1.7B as students, following the training settings in Section~5.
Figure~6 reports MATH-500 learning curves, with insets summarizing final accuracy.

\paragraph{Results.}
Figure~\ref{fig:beta-curves} shows (i) shorter initial stagnation as $\beta$ increases from 0 to 1 and (ii) non-monotonic final accuracy. For the 0.6B student, accuracy reaches 35.2\%, 44.6\%, and a peak of 44.8\% at $\beta=0,1,1.2$, respectively. For the 1.7B student, accuracy peaks at 61.8\% ($\beta=1$), versus 57.0\% ($\beta=0$) and 60.0\% ($\beta=1.2$). Appendix~\ref{app:probability-weighting-ablation} compares $\beta=0$ and $\beta=1$ across all seven benchmarks. \emph{Probability weighting improves learning speed and final accuracy over reward-only selection, but excessive weighting reduces performance: both students peak near $\beta=1$.}

\subsection{Main Results}
\label{sec:main-results}

\paragraph{Better performance with less supervision.}
Retaining only 40\% of response tokens, DIAL-OPD outperforms Vanilla OPD and both full-token variants in seven-benchmark mean accuracy across all four model pairs. Against Vanilla OPD, gains reach \textbf{5.25} percentage points in mean accuracy (4B$\to$0.6B, $\beta=1$) and \textbf{13.33} points on AIME25 Pass@16 (8B$\to$1.7B, $\beta=1.2$), doubling it from 13.33\% to 26.67\%. Against the strongest full-token variant, gains reach \textbf{1.24} points in mean accuracy (4B$\to$1.7B) and \textbf{10.00} points on AMC23 Pass@16 (8B$\to$1.7B), both at $\beta=1$ and over PowerOPD. \emph{These results challenge supervision density as a guiding principle for effective distillation: substantially less supervision can outperform standard and improved full-token OPD.}

\paragraph{Substantially outperforming existing token-selection methods.}
DIAL-OPD achieves the highest seven-benchmark mean accuracy across all eight combinations of model pair and retention ratio. At 20\% retention, it exceeds the strongest existing selection method by up to \textbf{5.23} percentage points in mean accuracy (4B$\to$1.7B, over CC-OPD) and \textbf{10.33} points on GSM8K Avg@4 (4B$\to$0.6B, over SA-OPD), both at $\beta=1$. \emph{Together, these comparisons demonstrate that DIAL-OPD identifies more effective supervision than existing token-selection methods at the same token budget.}

\paragraph{Effective supervision allocation can outweigh teacher scaling.}
DIAL-OPD with a 4B teacher outperforms full-token OPD with an 8B teacher at both student scales. With $\beta=1$ and 40\% retention, it achieves mean accuracies of 24.68\% and 35.36\% for the 0.6B and 1.7B students, respectively, exceeding not only Vanilla OPD but also the strongest full-token variant under the larger teacher by \textbf{0.95} and \textbf{1.76} percentage points. \emph{These findings suggest that allocating supervision effectively can yield greater gains than simply increasing teacher size.}

\subsection{Understanding DIAL-OPD's Selection Behavior}
\label{sec:selection-behavior}
\begin{figure}[t]
\centering
\includegraphics[width=\linewidth]{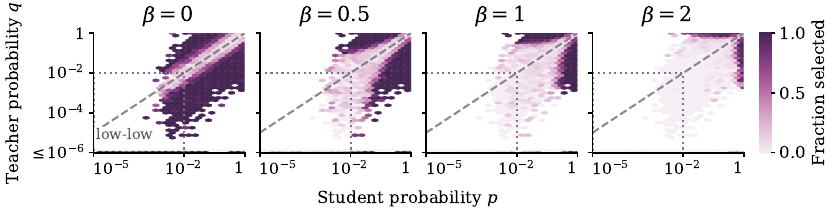}
\par\vspace{-10pt}
\caption{\small Token selection under varying $\beta$ at 20\% retention. Hexagon colors indicate selection rates; dotted lines delimit the low-low region, and teacher probabilities $q\leq10^{-6}$ are pooled for visualization.}
\label{fig:beta-selection-map}

\vspace{8pt}

\centering
\includegraphics[width=\linewidth]{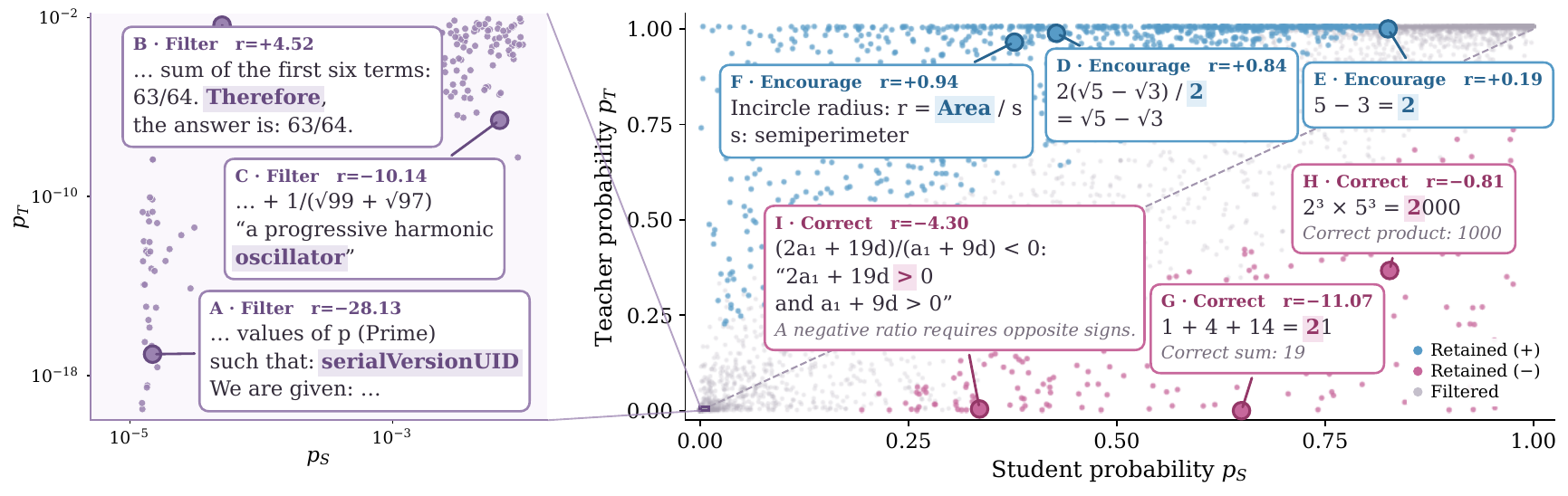}
\par\vspace{-10pt}
\caption{\small Token selections and signed OPD rewards in context ($\beta=1$, 20\% retention). \legendword{legendblue}{Blue}/\legendword{legendpink}{pink} mark retained tokens with positive/negative rewards; \legendword{legendgray}{gray} marks filtered tokens. Left: low-low zoom on log axes.}
\label{fig:token-selection-example}
\end{figure}

\paragraph{Moderate $\beta$ achieves the best balance in token selection.}
We examine how $\beta$, the dial between log-probability and probability spaces, shapes token selection. We compare $\beta=0,0.5,1,2$ on the same 128 responses to DeepScaleR prompts at 20\% retention. Figure~\ref{fig:beta-selection-map} reveals two stages: (i) increasing $\beta$ from 0 to 1 shifts selection away from the low-low region, reducing its share of selected tokens from 8.26\% to 0.29\%; (ii) increasing $\beta$ to 2 suppresses this share to 0.02\%, but also reduces selection in regions where one model assigns substantial probability while the other does not. Together with the performance trends in Section~\ref{sec:beta-ablation}, these observations suggest that \emph{moderate probability weighting improves performance by suppressing low-low disagreement, whereas excessive weighting may degrade performance by discarding useful signals supported by one model.}

\paragraph{Reward magnitude alone is insufficient to assess learning value.}
Figure~\ref{fig:token-selection-example} shows retained and filtered tokens in context ($\beta=1$, 20\% retention). DIAL-OPD filters the unrelated identifier \texttt{serialVersionUID} ($-28.13$) and \texttt{Therefore} ($+4.52$), a redundant transition to an already-obtained answer. Retained supervision instead encourages \texttt{Area} in a correct incircle formula ($+0.94$) and penalizes an incorrect arithmetic digit ($-0.81$). \emph{Tokens that do not advance reasoning can receive larger reward magnitudes than those providing useful supervision.} This contrast supports using both relative disagreement and absolute probability scale for selection.

\section{Conclusion}
\label{sec:conclusion}

We show that more supervision does not necessarily improve OPD, motivating selective supervision.
Yet prioritizing large teacher--student disagreements can overemphasize harmful low-low tokens.
DIAL-OPD incorporates absolute probability scale through the logarithmic mean to address this blind spot.
Selection analyses suggest that moderate probability weighting suppresses low-low tokens while preserving useful disagreements, whereas excessive weighting can discard valuable supervision.
Across four teacher--student pairs and seven benchmarks, it outperforms full-token OPD with fewer tokens and existing selection methods at matched retention ratios.
Effective supervision allocation can even outweigh teacher scaling, underscoring the importance of token selection.

\FloatBarrier
\bibliography{references}
\bibliographystyle{plainnat}

\appendix
\clearpage
\appendix
\section*{Appendix}
\section{Related Work}
\label{app:related-work}

\paragraph{On-policy distillation.}
OPD trains a student on its own generated responses using teacher supervision, reducing the mismatch between training and inference. GKD~\citep{agarwal2024gkd} studies flexible divergence objectives on student-generated sequences, MiniLLM~\citep{gu2024minillm} develops reverse-KL optimization, and \citet{lu2025opd} present a practical sampled-token training recipe. Subsequent developments address two complementary questions. \textbf{(i)} \emph{Constructing teacher supervision.} Qwen3~\citep{qwen2025qwen3} transfers capabilities from larger teachers through off-policy initialization followed by on-policy distillation; MiMo-V2-Flash~\citep{xiaomi2026mimov2flash} integrates domain-specialized teachers; GLM-5~\citep{glm2026glm5} recovers capabilities from earlier training checkpoints; and Kimi K3~\citep{kimi2026k3} consolidates experts specialized by domain and reasoning effort. Self-distillation instead constructs a teacher from the model itself: OPSD~\citep{zhao2026selfdistilled} provides it with solutions unavailable to the student, while SDPO~\citep{hubotter2026reinforcement} conditions it on textual feedback or successful attempts. \textbf{(ii)} \emph{Optimizing the distillation objective.} ExOPD~\citep{yang2026exopd} separates reward strength from KL regularization, while Direct-OPD~\citep{feng2026directopd} derives rewards from the teacher's policy change during reinforcement learning. PowerOPD~\citep{zhao2026poweropd} bounds rewards through power transformations, and Veto~\citep{jang2026stable} stabilizes alignment through a geometric target distribution combining teacher and student probabilities. AOPD~\citep{jia2026aopd} uses policy-gradient reinforcement for positive advantages and local distribution matching otherwise; EOPD~\citep{jin2026entropy} adds forward-KL supervision where teacher entropy is high. Recent evidence also highlights a bottleneck in supervision utilization: \citet{fu2026oneshot} show that training on a single query recovers much of full-data OPD's improvement while teacher--student alignment remains slow.

\paragraph{Selective supervision for OPD.}
Selective supervision addresses which parts of generated responses should contribute to learning by selecting, weighting, or restructuring teacher supervision. Two complementary approaches consider individual tokens and the structure of the surrounding trajectory. \textbf{(i)} \emph{Assessing token learning value.} TIP~\citep{xu2026tip} combines student uncertainty with teacher--student divergence, while REOPOLD~\citep{ko2026reopold} couples entropy-guided token sampling with reward clipping. SuRe~\citep{shao2026tokenanalysis} uses bounded surprise-based weighting to emphasize tokens assigned low probability by the student, and TA-OPD~\citep{wang2026taopd} assesses disagreement together with teacher support for the student's likely alternatives. Beyond distributional statistics, SA-OPD~\citep{jiang2026saopd} filters high-impact supervision weakly dependent on the question, whereas CROP~\citep{li2026crop} measures sensitivity to changed task conditions relative to meaning-preserving rewrites. USD~\citep{yang2026usd} jointly adjusts token weights and the information supplied to a self-teacher according to student learning capacity. TIDE~\citep{yu2026tide} applies different corrections to tokens overestimated by the student and alternatives favored by the teacher. \textbf{(ii)} \emph{Using trajectory structure.} Prefix OPD~\citep{zhang2026prefixopd} trains on truncated reasoning prefixes, and PG-OPD~\citep{zhao2026pgopd} uses teacher--student agreement on prefixes to decide which rollouts to extend. FiRe-OPD~\citep{li2026fireopd} filters trajectories by teacher likelihood before weighting their tokens, while TOPD~\citep{jiang2026topd} uses near-future continuations to guide corrections. DASH~\citep{hou2026dash} aggregates divergence across successive positions to construct token weights; ATOD~\citep{tan2026atod} weights interaction turns while gradually shifting from distillation to reinforcement learning. Evidence from extremely sparse OPD further shows that supervising one or two selected tokens per trajectory can match or exceed dense training~\citep{liu2026sparse}. Building on these findings, we identify how relative disagreement can overemphasize tokens assigned low probability by both models and test their training value through targeted replacement. \method{} uses the logarithmic mean to connect relative disagreement with absolute probability scale, selecting a fixed fraction of tokens within each response while preserving the original signed OPD reward.

\section{Experimental Details}
\label{app:experimental-details}

\paragraph{Models.}
Students are initialized from Qwen3-0.6B-Base and Qwen3-1.7B-Base, while teachers use the instruction-tuned Qwen3-4B and Qwen3-8B checkpoints. The equal-size comparison in Section~\ref{sec:analysis-supervision} additionally uses Qwen3-4B-Base as the student. We omit the ``-Base'' suffix in the main text for brevity.

\paragraph{Training details.}
All methods use AdamW with BF16 mixed precision, learning rate $5\times10^{-7}$, and batch size 32. Each rollout contains up to 1,024 response tokens. Training rollouts use temperature 1.0, top-$p$ 0.95, and top-$k$ 20. Prompt, padding, and terminal EOS positions are excluded from the training loss for all methods.

\paragraph{Evaluation protocol.}
For the main comparisons, we sample $k=4$ responses per question on GSM8K, MATH-500, and OlympiadBench; $k=8$ on Minerva Math; and $k=16$ on AMC23, AIME24, and AIME25. Sampling uses temperature 0.6 and top-$p$ 0.95. Tables abbreviate average correctness, success in at least one sampled response, and mean response length as Avg, Pass, and Len, respectively. Mean denotes the unweighted average of Avg across the seven benchmarks.

\subsection{Supervision Density Experiment}
\label{app:supervision-density}

Training follows the configuration in Section~\ref{sec:experimental-setup}. For each response, we exclude prompt, padding, and terminal EOS positions, leaving $N$ eligible tokens. We uniformly sample $K=\max(1,\operatorname{round}(\rho N))$ positions without replacement, where $\rho$ is the retention ratio. Selection masks are refreshed at every training step. Only selected positions contribute to the original signed OPD loss, averaged over all selected tokens in the batch; unselected tokens remain in the context. We evaluate on MATH-500 before training and every 50 steps, sampling one response per question with temperature 0.6, top-$p$ 0.95, and a maximum response length of 4,096 tokens. The same-size experiment repeats the retention-ratio sweep with Qwen3-4B-Base as the student and Qwen3-4B as the teacher.

\subsection{Baseline Implementations}
\label{app:baseline-implementations}

We adapt token-selection baselines to matched per-response retention ratios. For SA-OPD, we select tokens by input-dependence ranking instead of its original joint filtering conditions. For TA-OPD, teacher support measures the probability mass that the teacher assigns to the student's top-$K$ candidate tokens~\citep{wang2026taopd}; we retain batch-level normalization and perform selection within each response. Prefix OPD selects the initial fraction of each complete rollout rather than stopping generation early. For CC-OPD, the teacher's pre-RL checkpoint is unavailable, so we use a smaller instruction-tuned model from the same family as the reference and select tokens with the largest absolute teacher--reference probability gaps.

\subsection{Low-Low Token Analysis}
\label{app:low-low-analysis}

For the concentration analysis in Section~\ref{sec:analysis-low-low}, we pool selected-token counts and absolute reward sums across responses before computing low-low shares, rather than averaging per-response percentages. The token share divides the selected low-low count by the total selected count; the reward share divides their summed absolute rewards by the absolute reward sum over all selected tokens. Replacement candidates are ranked by absolute reward, while the OPD loss retains the original reward signs. Both replacement strategies use training and evaluation settings matched to the corresponding Reward-Top baseline at each retention ratio.

\section{Derivation of the Sampled-Token OPD Update}
\label{app:opd-update-derivation}
\setcounter{equation}{0}
\renewcommand{\theequation}{\thesection.\arabic{equation}}

We derive the update in Eq.~\eqref{eq:opd-update} by differentiating the conditional reverse KL while holding the rollout context distribution fixed. The derivation also establishes the unbiasedness of its Monte Carlo estimator and explains the stop-gradient treatment of the token-level reward.

\paragraph{Fixed-context objective.}
Let $\bar\theta$ denote the student parameters used to collect rollouts, and let $\theta$ denote the parameters with respect to which we differentiate. For $x\sim\mathcal D$ and $y\sim\pi_{\bar\theta}(\cdot\mid x)$, the context preceding token $y_t$ is $c_t=(x,y_{<t})$. With the teacher policy fixed, define
\begin{equation}
\mathcal L_{\mathrm{ctx}}(\theta;\bar\theta)
=\mathbb E_{x\sim\mathcal D,\;y\sim\pi_{\bar\theta}(\cdot\mid x)}
\left[\sum_{t=1}^{|y|}
D_{\mathrm{KL}}\!\left(\pi_\theta(\cdot\mid c_t)\,\middle\|\,\pi_T(\cdot\mid c_t)\right)\right].
\label{eq:app-fixed-context}
\end{equation}
Here, $\bar\theta$ is held fixed during differentiation: the contexts come from the rollout policy, while the conditional student distributions at those contexts depend on $\theta$. We evaluate the resulting gradient at $\theta=\bar\theta$. We assume differentiable, strictly positive student and teacher probabilities over a finite vocabulary, bounded rollout lengths, and the regularity conditions required to interchange differentiation and expectation.

\paragraph{From reverse KL to a policy-gradient update.}
For a fixed context $c$, write $p_\theta(v)=\pi_\theta(v\mid c)$ and $q(v)=\pi_T(v\mid c)$. Differentiating the vocabulary-wide reverse KL gives
\begin{equation}
\begin{aligned}
\nabla_\theta D_{\mathrm{KL}}(p_\theta\|q)
&=\nabla_\theta\sum_{v\in\mathcal V}p_\theta(v)\log\frac{p_\theta(v)}{q(v)}\\
&=\sum_{v\in\mathcal V}\left(\log\frac{p_\theta(v)}{q(v)}+1\right)\nabla_\theta p_\theta(v)\\
&=\sum_{v\in\mathcal V}p_\theta(v)\log\frac{p_\theta(v)}{q(v)}\,\nabla_\theta\log p_\theta(v).
\end{aligned}
\label{eq:app-kl-gradient}
\end{equation}
The last equality uses $\nabla_\theta p_\theta(v)=p_\theta(v)\nabla_\theta\log p_\theta(v)$ and the normalization identity
\begin{equation}
\sum_{v\in\mathcal V}\nabla_\theta p_\theta(v)
=\nabla_\theta\sum_{v\in\mathcal V}p_\theta(v)
=\nabla_\theta 1=0.
\label{eq:app-normalization}
\end{equation}
Consequently, the negative gradient is an expectation of the teacher--student log-ratio multiplied by the student's log-probability gradient. Averaging this identity over the fixed rollout context distribution and evaluating at the rollout policy yields
\begin{equation}
\begin{aligned}
g_{\mathrm{OPD}}(\bar\theta)
&:=-\left.\nabla_\theta\mathcal L_{\mathrm{ctx}}(\theta;\bar\theta)\right|_{\theta=\bar\theta}\\
&=\mathbb E_{x\sim\mathcal D,\;y\sim\pi_{\bar\theta}(\cdot\mid x)}
\left[\sum_{t=1}^{|y|}
\underbrace{\log\frac{\pi_T(y_t\mid c_t)}{\pi_{\bar\theta}(y_t\mid c_t)}}_{r_t^{\mathrm{OPD}}}
\left.\nabla_\theta\log\pi_\theta(y_t\mid c_t)\right|_{\theta=\bar\theta}\right].
\end{aligned}
\label{eq:app-opd-direction}
\end{equation}
This is the REINFORCE policy-gradient form, with the log-ratio serving as the token-level reward~\citep{williams1992reinforce,sutton1999policygradient}. Removing the distinction between the differentiation variable and the rollout parameters after evaluation recovers the notation of Eq.~\eqref{eq:opd-update}.

\paragraph{Monte Carlo estimation and stop-gradient implementation.}
Given $B$ independent prompt--response pairs sampled from $\mathcal D$ and $\pi_{\bar\theta}$, let $c_{i,t}$ denote the context of token $y_{i,t}$, and let $r_{i,t}^{\mathrm{OPD}}$ be its log-ratio reward evaluated at $\bar\theta$. The sampled update is
\begin{equation}
\widehat g_B=\frac{1}{B}\sum_{i=1}^{B}\sum_{t=1}^{|y_i|}
r_{i,t}^{\mathrm{OPD}}
\left.\nabla_\theta\log\pi_\theta(y_{i,t}\mid c_{i,t})\right|_{\theta=\bar\theta},
\qquad\mathbb E[\widehat g_B]=g_{\mathrm{OPD}}(\bar\theta).
\label{eq:app-mc-update}
\end{equation}
Each sampled token supplies one Monte Carlo sample of the conditional expectation in Eq.~\eqref{eq:app-kl-gradient}. Tokens within a response need not be independent: conditioned on its prefix, each token is sampled from the student distribution, and unbiasedness follows from conditional expectation and linearity. In implementation, the same update is obtained by minimizing
\begin{equation}
\widehat\ell_{\mathrm{OPD}}(\theta;\bar\theta)
=-\frac{1}{B}\sum_{i=1}^{B}\sum_{t=1}^{|y_i|}
\operatorname{sg}\!\left[r_{i,t}^{\mathrm{OPD}}\right]
\log\pi_\theta(y_{i,t}\mid c_{i,t}),
\label{eq:app-surrogate}
\end{equation}
where $\operatorname{sg}$ denotes the stop-gradient operator. By construction,
\begin{equation}
-\left.\nabla_\theta\widehat\ell_{\mathrm{OPD}}(\theta;\bar\theta)\right|_{\theta=\bar\theta}
=\widehat g_B.
\label{eq:app-surrogate-gradient}
\end{equation}
The stop-gradient operation implements the coefficient obtained from the analytical KL gradient; it does not assume that the original log-ratio is independent of the student parameters. Its explicit parameter dependence has already been accounted for in Eq.~\eqref{eq:app-kl-gradient}, where the additional term cancels by normalization. Thus, Eq.~\eqref{eq:app-surrogate} is a surrogate loss that reproduces the required gradient at the rollout policy, rather than a direct differentiation of the sampled log-ratio alone.

\paragraph{Relation to the full on-policy objective.}
At $\theta=\bar\theta$, the fixed-context objective in Eq.~\eqref{eq:app-fixed-context} has the same value as the objective in Eq.~\eqref{eq:opd-objective}, but their derivatives need not coincide. Differentiating Eq.~\eqref{eq:opd-objective} through the entire trajectory distribution additionally accounts for how earlier token choices change the contexts encountered later in the response. Equation~\eqref{eq:opd-update} instead uses the conditional reverse-KL gradients at the sampled contexts, following the immediate-token update used in on-policy distillation~\citep{lu2025opd}. The unbiasedness statement in Eq.~\eqref{eq:app-mc-update} therefore concerns this fixed-context update direction. Contexts are nevertheless refreshed by new student rollouts during training; holding their distribution fixed during differentiation does not make the training procedure off-policy.

\section{Probability Weighting Ablation}
\label{app:probability-weighting-ablation}

To isolate the contribution of probability weighting, Table~\ref{tab:probability-weighting-ablation} compares DIAL-OPD at $\beta=0$ and $\beta=1$ under the same retention ratio across all four teacher--student pairs. The $\beta=0$ setting corresponds to Reward-Top in Section~\ref{sec:analysis-reward}. With a Qwen3-4B teacher and Qwen3-1.7B-Base student at 40\% retention, $\beta=1$ improves seven-benchmark mean accuracy by \textbf{6.96} percentage points and GSM8K Avg@4 by \textbf{15.96} points over $\beta=0$. These results extend the learning-dynamics analysis in Section~\ref{sec:beta-ablation} to the full evaluation suite.

\begin{table}[t]
\centering
\caption{Probability-weighting ablation across four teacher--student pairs and seven benchmarks. Each retention ratio compares reward-only selection ($\beta=0$) with $\beta=1$; \textbf{bold} marks the higher accuracy within each pair, including ties.}
\label{tab:probability-weighting-ablation}
\begingroup
\fontsize{6.7}{8.4}\selectfont
\setlength{\tabcolsep}{1.65pt}
\renewcommand{\arraystretch}{1.00}
\setlength{\aboverulesep}{1.6pt}
\setlength{\belowrulesep}{1.6pt}
\sbox{\resultstablebox}{%
\begin{tabular}{@{}c@{\hspace{2.5pt}}l@{\hspace{2.5pt}}*{7}{rr>{\color{black!64}}r@{\hspace{4pt}}}r@{}}
\toprule
Keep & $\beta$ & \multicolumn{3}{c}{GSM8K} & \multicolumn{3}{c}{MATH500} & \multicolumn{3}{c}{AMC23} & \multicolumn{3}{c}{AIME24} & \multicolumn{3}{c}{AIME25} & \multicolumn{3}{c}{Minerva} & \multicolumn{3}{c}{Olympiad} & Mean \\
\cmidrule(lr){3-5} \cmidrule(lr){6-8} \cmidrule(lr){9-11} \cmidrule(lr){12-14} \cmidrule(lr){15-17} \cmidrule(lr){18-20} \cmidrule(lr){21-23}
 & & Avg & Pass & Len & Avg & Pass & Len & Avg & Pass & Len & Avg & Pass & Len & Avg & Pass & Len & Avg & Pass & Len & Avg & Pass & Len & Avg \\
\midrule
\rowcolor{studentband}
\multicolumn{24}{c}{\rule{0pt}{9.5pt}\textbf{Teacher: Qwen3-4B $\to$ Student: Qwen3-0.6B-Base}\rule[-2pt]{0pt}{0pt}} \\
\midrule
20\% & 0 & 57.3 & 79.2 & 535 & \textbf{36.4} & \textbf{52.4} & 1786 & \textbf{15.9} & \textbf{57.5} & 5254 & \textbf{1.7} & \textbf{10.0} & 7297 & 0.8 & 6.7 & 7685 & \textbf{8.5} & \textbf{20.6} & 1593 & 12.9 & \textbf{24.6} & 5419 & 19.1 \\
\rowcolor{methodblue}
\cellcolor{white} & 1 & \textbf{67.0} & \textbf{81.3} & 1072 & 34.2 & 47.2 & 2609 & \textbf{15.9} & 45.0 & 6554 & 0.4 & 6.7 & 7985 & \textbf{1.5} & \textbf{10.0} & 8016 & 7.6 & 16.9 & 2617 & \textbf{13.1} & 24.2 & 6658 & \textbf{19.9} \\
\cmidrule(lr){2-24}
40\% & 0 & 56.1 & 81.3 & 507 & 34.1 & 52.4 & 1752 & 15.8 & \textbf{52.5} & 4996 & 1.9 & 3.3 & 7180 & 0.4 & \textbf{6.7} & 7516 & 8.0 & 21.0 & 1504 & 12.4 & 24.5 & 5495 & 18.4 \\
\rowcolor{methodblue}
\cellcolor{white} & 1 & \textbf{69.1} & \textbf{84.7} & 578 & \textbf{44.9} & \textbf{61.6} & 1714 & \textbf{25.5} & 50.0 & 4774 & \textbf{2.5} & \textbf{13.3} & 7001 & \textbf{1.2} & \textbf{6.7} & 7179 & \textbf{10.7} & \textbf{22.1} & 1484 & \textbf{18.8} & \textbf{33.5} & 4705 & \textbf{24.7} \\
\midrule
\rowcolor{studentband}
\multicolumn{24}{c}{\rule{0pt}{9.5pt}\textbf{Teacher: Qwen3-4B $\to$ Student: Qwen3-1.7B-Base}\rule[-2pt]{0pt}{0pt}} \\
\midrule
20\% & 0 & 69.4 & 90.8 & 354 & 53.1 & \textbf{71.6} & 1149 & 30.0 & \textbf{75.0} & 3476 & 6.0 & 20.0 & 5589 & 4.2 & \textbf{16.7} & 5478 & 14.7 & 29.4 & 943 & 24.0 & 39.2 & 3568 & 28.8 \\
\rowcolor{methodblue}
\cellcolor{white} & 1 & \textbf{82.3} & \textbf{91.8} & 404 & \textbf{57.8} & 70.2 & 1445 & \textbf{37.8} & \textbf{75.0} & 4276 & \textbf{8.8} & \textbf{26.7} & 6287 & \textbf{4.8} & \textbf{16.7} & 6170 & \textbf{17.9} & \textbf{32.7} & 1216 & \textbf{29.5} & \textbf{43.2} & 4222 & \textbf{34.1} \\
\cmidrule(lr){2-24}
40\% & 0 & 67.2 & 91.9 & 363 & 51.6 & 72.0 & 1124 & 27.7 & \textbf{82.5} & 3272 & 8.3 & \textbf{30.0} & 5530 & \textbf{5.4} & 13.3 & 5303 & 14.3 & 29.0 & 872 & 24.3 & 41.1 & 3393 & 28.4 \\
\rowcolor{methodblue}
\cellcolor{white} & 1 & \textbf{83.2} & \textbf{92.8} & 344 & \textbf{61.5} & \textbf{73.0} & 1223 & \textbf{38.8} & 75.0 & 3457 & \textbf{9.8} & 26.7 & 5992 & 5.2 & \textbf{16.7} & 5951 & \textbf{19.4} & \textbf{30.9} & 992 & \textbf{29.7} & \textbf{44.5} & 3789 & \textbf{35.4} \\
\midrule
\rowcolor{studentband}
\multicolumn{24}{c}{\rule{0pt}{9.5pt}\textbf{Teacher: Qwen3-8B $\to$ Student: Qwen3-0.6B-Base}\rule[-2pt]{0pt}{0pt}} \\
\midrule
20\% & 0 & 58.2 & 80.4 & 471 & 39.2 & 57.4 & 1568 & 15.8 & 57.5 & 4527 & 1.5 & 3.3 & 6626 & 0.2 & 3.3 & 7127 & 8.0 & 20.6 & 1402 & 13.8 & 25.8 & 5025 & 19.5 \\
\rowcolor{methodblue}
\cellcolor{white} & 1 & \textbf{63.8} & \textbf{83.2} & 465 & \textbf{42.3} & \textbf{59.6} & 1633 & \textbf{21.2} & \textbf{62.5} & 4372 & \textbf{1.7} & \textbf{10.0} & 6566 & \textbf{2.1} & \textbf{10.0} & 6583 & \textbf{9.7} & \textbf{23.2} & 1265 & \textbf{16.0} & \textbf{29.1} & 4494 & \textbf{22.4} \\
\cmidrule(lr){2-24}
40\% & 0 & 58.8 & 80.5 & 488 & 40.9 & 57.6 & 1562 & 16.9 & 57.5 & 4697 & 1.9 & \textbf{13.3} & 6521 & \textbf{1.2} & \textbf{10.0} & 7080 & 9.9 & \textbf{22.8} & 1276 & 14.4 & 27.2 & 4967 & 20.6 \\
\rowcolor{methodblue}
\cellcolor{white} & 1 & \textbf{66.6} & \textbf{84.6} & 436 & \textbf{45.6} & \textbf{61.4} & 1482 & \textbf{27.3} & \textbf{62.5} & 3872 & \textbf{3.1} & \textbf{13.3} & 6432 & \textbf{1.2} & 6.7 & 6589 & \textbf{10.6} & 22.1 & 1197 & \textbf{17.7} & \textbf{29.7} & 4392 & \textbf{24.6} \\
\midrule
\rowcolor{studentband}
\multicolumn{24}{c}{\rule{0pt}{9.5pt}\textbf{Teacher: Qwen3-8B $\to$ Student: Qwen3-1.7B-Base}\rule[-2pt]{0pt}{0pt}} \\
\midrule
20\% & 0 & 74.8 & 91.7 & 345 & 58.2 & 73.8 & 1116 & 31.4 & \textbf{72.5} & 3309 & 7.5 & \textbf{16.7} & 6003 & \textbf{6.5} & \textbf{16.7} & 5179 & 16.3 & 30.1 & 931 & 27.0 & 42.0 & 3494 & 31.7 \\
\rowcolor{methodblue}
\cellcolor{white} & 1 & \textbf{80.6} & \textbf{92.9} & 333 & \textbf{60.2} & \textbf{74.4} & 1199 & \textbf{35.8} & 70.0 & 3404 & \textbf{7.9} & \textbf{16.7} & 5791 & 5.4 & \textbf{16.7} & 5339 & \textbf{18.3} & \textbf{31.2} & 967 & \textbf{28.2} & \textbf{42.3} & 3570 & \textbf{33.8} \\
\cmidrule(lr){2-24}
40\% & 0 & 80.0 & 92.4 & 347 & 58.1 & 71.8 & 1183 & 33.1 & 80.0 & 3262 & 7.3 & \textbf{20.0} & 5828 & 5.6 & 20.0 & 5390 & 17.5 & \textbf{32.7} & 991 & 26.5 & 40.2 & 3693 & 32.6 \\
\rowcolor{methodblue}
\cellcolor{white} & 1 & \textbf{82.4} & \textbf{93.5} & 323 & \textbf{59.7} & \textbf{73.8} & 1173 & \textbf{38.0} & \textbf{82.5} & 3098 & \textbf{8.8} & 16.7 & 5764 & \textbf{6.5} & \textbf{23.3} & 5425 & \textbf{18.0} & 31.6 & 913 & \textbf{28.9} & \textbf{43.5} & 3533 & \textbf{34.6} \\
\bottomrule
\end{tabular}}
\typeout{STACKED-WIDTH=\the\wd\resultstablebox; LIMIT=\the\linewidth}
\typeout{STACKED-HEIGHT=\the\dimexpr\ht\resultstablebox+\dp\resultstablebox\relax; TEXTHEIGHT=\the\textheight}
\usebox{\resultstablebox}
\endgroup
\end{table}

\section{Proofs of DIAL-OPD Properties}
\label{app:dial-proofs}

\subsection{Proof of Proposition~\ref{prop:low-low-scaling}}
\label{app:proposition-one-proof}

Fix $p,q\in(0,1]$, $\beta\geq0$, and $\varepsilon\in(0,1]$. Scaling both probabilities by $\varepsilon$ preserves their ratio while scaling their logarithmic mean by the same factor. From the integral definition in Equation~\ref{eq:logarithmic-mean},
\begin{equation}
\begin{aligned}
L(\varepsilon p,\varepsilon q)
&=\int_0^1(\varepsilon p)^{1-u}(\varepsilon q)^u\,du\\
&=\int_0^1\varepsilon^{1-u+u}p^{1-u}q^u\,du
=\varepsilon L(p,q).
\end{aligned}
\end{equation}
Meanwhile, $\left|\log[(\varepsilon q)/(\varepsilon p)]\right|=|\log(q/p)|$. Substituting these identities into Equation~\ref{eq:dial-score} yields
\begin{equation}
\begin{aligned}
s_\beta(\varepsilon p,\varepsilon q)
&=L(\varepsilon p,\varepsilon q)^\beta
\left|\log\frac{\varepsilon q}{\varepsilon p}\right|\\
&=[\varepsilon L(p,q)]^\beta\left|\log\frac{q}{p}\right|
=\varepsilon^\beta s_\beta(p,q).
\end{aligned}
\end{equation}
This identity also holds when $p=q$, since both scores are zero. For $p\ne q$, the score is positive; hence $\varepsilon<1$ leaves it unchanged at $\beta=0$ and strictly reduces it at $\beta>0$. Larger $\beta$ produces a smaller scaling factor $\varepsilon^\beta$, establishing stronger suppression under the same proportional reduction in both probabilities.

\subsection{Proof of Proposition~\ref{prop:one-sided-low-probabilities}}
\label{app:proposition-two-proof}

Fix $a\in(0,1]$ and $\beta\geq0$, and let $b\to0^+$. We prove the limit of the selection score from Equation~\ref{eq:dial-score} when one probability vanishes and the other remains fixed. Since $a>0$, we may restrict attention to $0<b<a$ without affecting the limit. Both probabilities are strictly positive throughout the derivation; the score need not be defined at $b=0$.

\paragraph{Symmetry.}
The logarithmic mean and the absolute log-ratio are symmetric in their arguments. In particular,
\begin{equation}
L(a,b)=\frac{a-b}{\log a-\log b}
=\frac{b-a}{\log b-\log a}=L(b,a),
\qquad
\left|\log\frac{a}{b}\right|
=\left|\log\frac{b}{a}\right|.
\end{equation}
It follows that $s_\beta(a,b)=s_\beta(b,a)$ for every positive $a,b$. Thus, the same limit applies whether the student probability or the teacher probability approaches zero, and it suffices to analyze $s_\beta(a,b)$.

\paragraph{Asymptotic form.}
For $0<b<a$, both $a-b$ and $\log(a/b)$ are positive. Substituting the closed form of the logarithmic mean into the score gives the exact identity
\begin{equation}
\begin{aligned}
s_\beta(a,b)
&=L(a,b)^\beta\left|\log\frac{a}{b}\right|\\
&=\left(\frac{a-b}{\log(a/b)}\right)^\beta\log(a/b)\\
&=(a-b)^\beta[\log(a/b)]^{1-\beta}.
\end{aligned}
\end{equation}
As $b\to0^+$, the factor $(a-b)^\beta$ converges to $a^\beta>0$, while $\log(a/b)=\log a-\log b\to+\infty$. More precisely,
\begin{equation}
\frac{s_\beta(a,b)}{a^\beta[\log(a/b)]^{1-\beta}}
=\left(1-\frac{b}{a}\right)^\beta\longrightarrow1,
\qquad
s_\beta(a,b)\sim a^\beta[\log(a/b)]^{1-\beta}.
\end{equation}
Thus, the limiting behavior is determined by the exponent $1-\beta$, since the remaining factor approaches a finite, strictly positive constant.

\paragraph{The three regimes.}
(i) If $0\leq\beta<1$, then $1-\beta>0$, so $[\log(a/b)]^{1-\beta}\to+\infty$ and hence $s_\beta(a,b)\to+\infty$. (ii) If $\beta=1$, the logarithmic factor cancels exactly: $s_1(a,b)=a-b\to a$. (iii) If $\beta>1$, then
\begin{equation}
s_\beta(a,b)=\frac{(a-b)^\beta}{[\log(a/b)]^{\beta-1}}\longrightarrow0,
\end{equation}
because the numerator converges to $a^\beta$ while the denominator diverges. Combining these cases with symmetry establishes both limits in Proposition~\ref{prop:one-sided-low-probabilities}. In particular, the zero limit for $\beta>1$ concerns arbitrarily small positive $b$; it does not imply that the score equals zero at any finite $b\in(0,a)$.

\end{document}